\documentclass[runningheads]{llncs}
\usepackage[numbers]{natbib}
\usepackage{graphicx}
\usepackage{pdfpages}
\usepackage{graphicx}
\usepackage{tikz}
\usepackage{comment}
\usepackage{amsmath,amssymb} 
\usepackage{color}
\usepackage{caption}
\usepackage{subcaption}
\usepackage{booktabs}

\usepackage[accsupp]{axessibility}  

\usepackage[pagebackref,breaklinks,colorlinks]{hyperref}

\usepackage{xspace}
\newcommand{\smpl}{SMPL\xspace}
\newcommand{\ghum}{GHUM\xspace}
\newcommand{\ghumHead}{GHUM-Head\xspace}
\newcommand{\ghumHand}{GHUM-Hand\xspace}

\newcommand{\supmat}{Supp.~Mat.\xspace}

\newcommand{\modelname}{SUPR\xspace}
\newcommand{\modelnameHead}{SUPR-Head\xspace}
\newcommand{\modelnameHand}{SUPR-Hand\xspace}
\newcommand{\modelnameFoot}{SUPR-Foot\xspace}

\newcommand{\smplx}{SMPL-X\xspace}
\newcommand{\mano}{MANO\xspace}
\newcommand{\flame}{FLAME\xspace}

\begin{document}
\bibliographystyle{unsrt}

\pagestyle{headings}
\mainmatter

\title{SUPR: A Sparse Unified Part-Based Human Representation} 

\titlerunning{SUPR}
\author{Ahmed A. A. Osman \inst{1} \and
Timo Bolkart \inst{1} \and
Dimitrios Tzionas\inst{2} 
\and  \\ 
Michael J. Black\inst{1} }
\authorrunning{Osman et al.}
\institute{Max Planck Institute for Intelligent Systems, T{\"u}bingen, Germany  \and 
University of Amsterdam\\
\email{\{aosman,tbolkart,black\}@tuebingen.mpg.de,d.tzionas@uva.nl}}
\maketitle

\begin{abstract}
Statistical 3D shape models of the head, hands, and full body are widely used in computer vision and graphics.
Despite their wide use, we show that existing models of the head and hands fail to capture the full range of motion for these parts.
Moreover, existing work largely ignores the feet, which are crucial for modeling human movement and have applications in biomechanics, animation, and the footwear industry.
The problem is that previous body part models are trained using 3D scans that are isolated to the individual parts.
Such data does not capture the full range of motion for such parts, e.g.~the motion of head relative to the neck.
Our observation is that full-body scans provide important information about the motion of the body parts.
Consequently, we propose a new  learning scheme that jointly trains a full-body model and specific part models using a federated dataset of full-body and body-part scans. Specifically, we train an expressive human body model called \modelname (Sparse Unified Part-Based Representation), where each joint strictly influences a sparse set of model vertices. 
The factorized representation enables separating \modelname into an entire suite of body part models: an expressive head (\modelnameHead), an articulated hand (\modelnameHand), and a novel foot (\modelnameFoot). 
Note that feet have received little attention and existing 3D body models have highly under-actuated feet.
Using novel 4D scans of feet, we train a model with an extended kinematic tree that captures the range of motion of the toes.
Additionally, feet deform due to ground contact.
To model this, we include a novel non-linear deformation function that predicts foot deformation conditioned on the foot pose, shape, and ground contact. 
We train \modelname on  an unprecedented number of scans: $1.2$ million body, head, hand and foot scans. 
We quantitatively compare \modelname and the separate body parts to existing expressive human body models and body-part models and find that our suite of models generalizes better and captures the body parts' full range of motion. \modelname is publicly available for research purposes at \href{http://supr.is.tue.mpg.de}{http://supr.is.tue.mpg.de} 
\end{abstract}

\section{Introduction}
\begin{figure}
    \includegraphics[width=\textwidth,trim={0 0 0 10cm},clip]{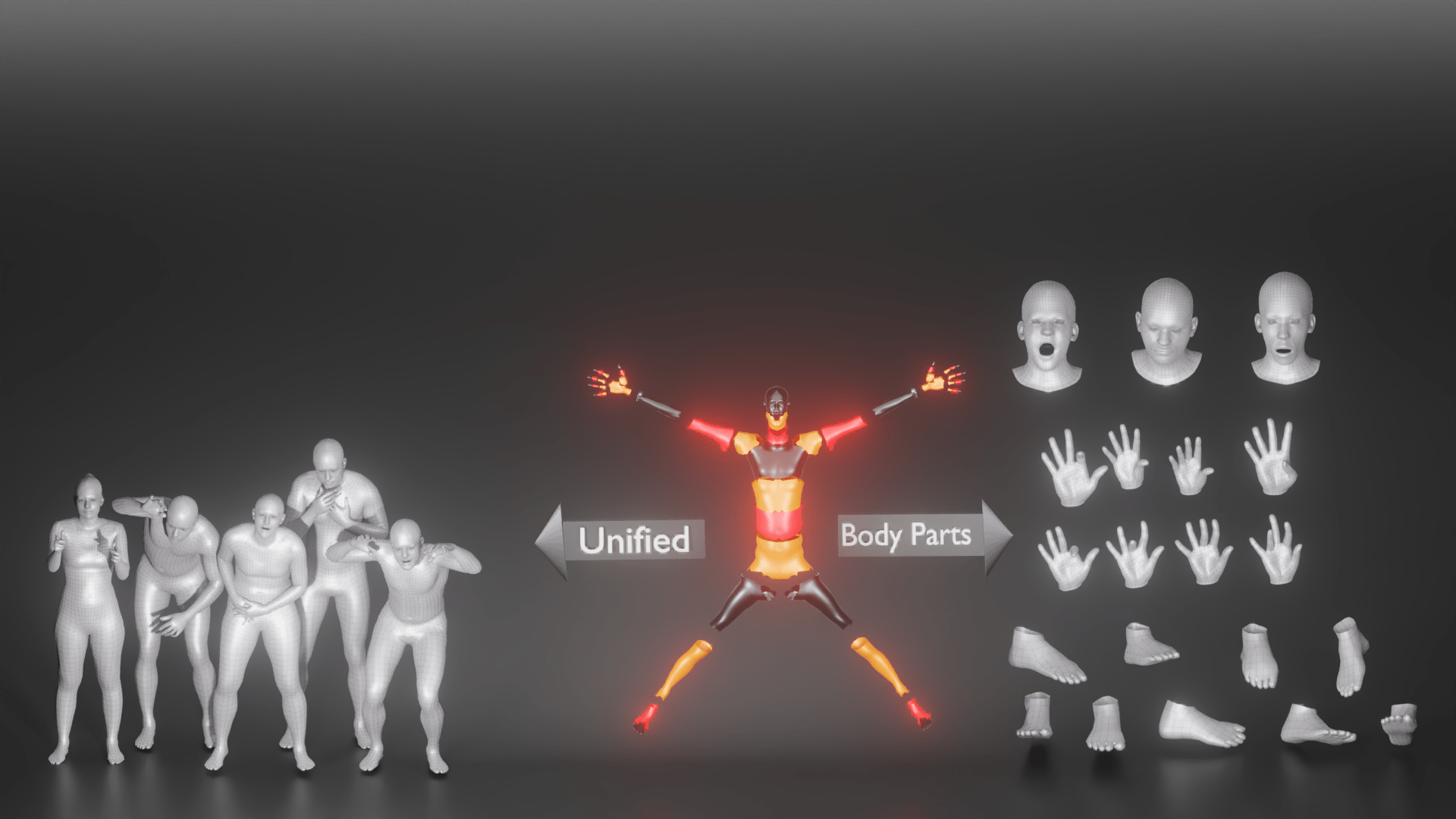}
    \caption{\textbf{Expressive part-based human body model.} \modelname is a factorized representation of the human body that can be separated into a full suite of body part models.}
    \label{fig:teaser}
\end{figure}

Generative 3D models of the human body and its parts play an important role in understanding human behaviour. 
Over the past two decades, numerous 3D models of the body \cite{Allen:2003,Anguelov05,chen2013tensor,Hasler2009,Hirshberg2012,SMPL:2015,STAR:ECCV:2020,Pishchulin2017,Wang_2020_ECCV}, face \cite{Amberg2008,Brunton2014,cao2014facewarehouse,FLAME:SiggraphAsia2017,li2020learning,COMA:ECCV18,yang2020facescape,Vlasic2005} and hands \cite{khamis2015learning,Kulon2019,Oikonomidis2011,MANO:SIGGRAPHASIA:2017,Smith2020,Tkach2016} have been proposed. 
Such models enabled a myriad of applications ranging from reconstructing bodies~\cite{kanazawa2018end,kocabas2019vibe,SPIN:ICCV:2019}, faces~\cite{Feng2018,Tewari2019,RingNet:CVPR:2019}, and hands \cite{Boukhayma2019,Hasson:CVPR:2019} from images and videos, modeling human interactions~\cite{fieraru2020three}, generating 3D clothed humans~\cite{alldieck2019tex2shape,Lassner:GP:2017,CAPE:CVPR:20,shape_under_cloth:CVPR17,Pons-Moll:Siggraph2017,bhatnagar2019mgn,patel20tailornet}, or generating humans in scenes~\cite{zanfir2020human,zhang2020generating,PLACE:3DV:2020}.
They are also used as priors for fitting models to a wide range of sensory input measurements like motion capture markers~\cite{Loper:SIGASIA:2014,AMASS:ICCV:2019} or IMUs~\cite{vonmarcardponsmollPAMI16,MuVS:3DV:2017,DIP:SIGGRAPHAsia:2018}.

\begin{figure}[t]
 \begin{subfigure}[b]{0.5\linewidth}
 \includegraphics[width=\textwidth]{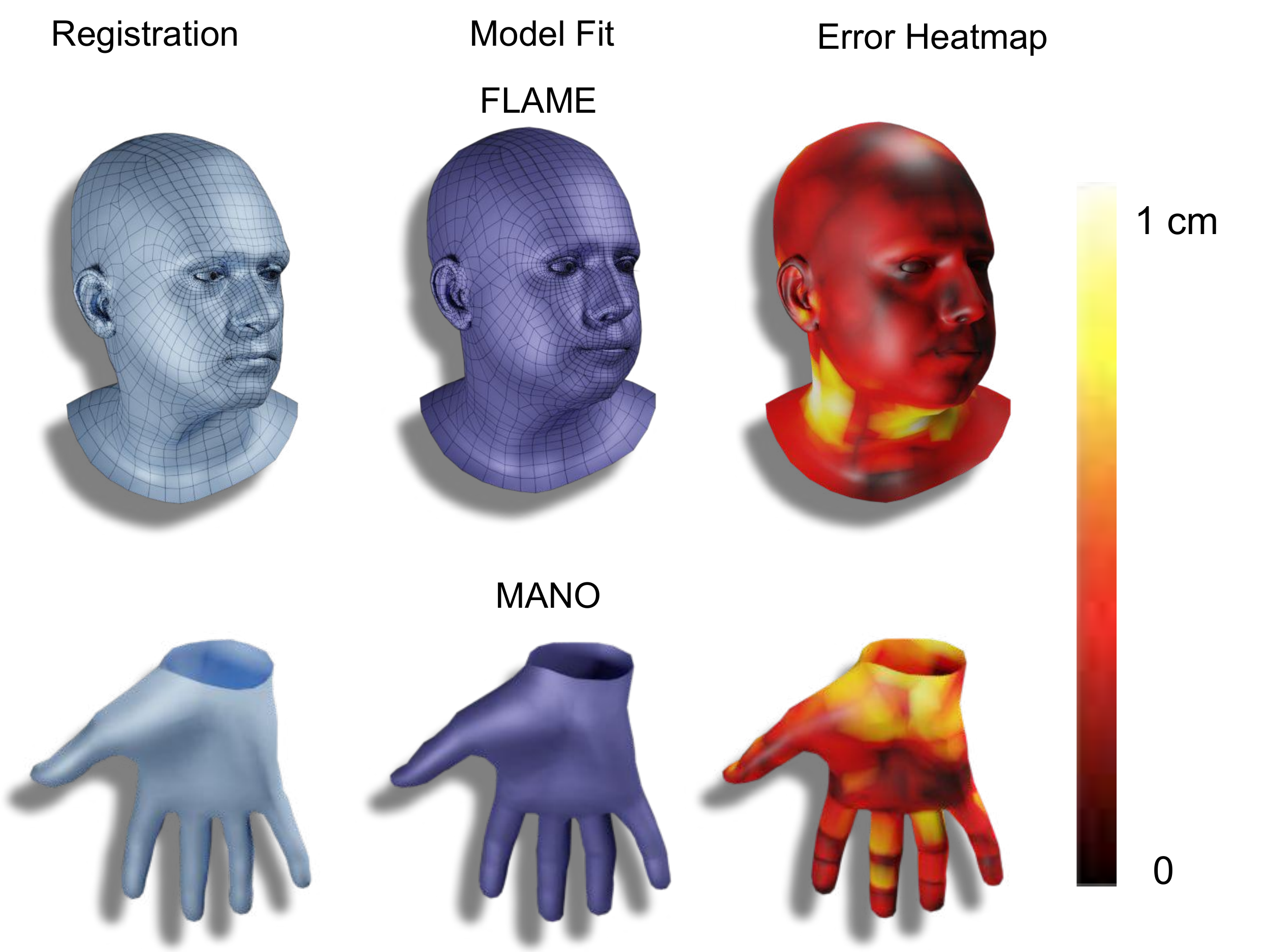}
 \caption{Body part models boundary error. }
 \label{fig:flame_mano_problems}
\end{subfigure}
 \begin{subfigure}[b]{0.5\linewidth}
 \includegraphics[width=\textwidth]{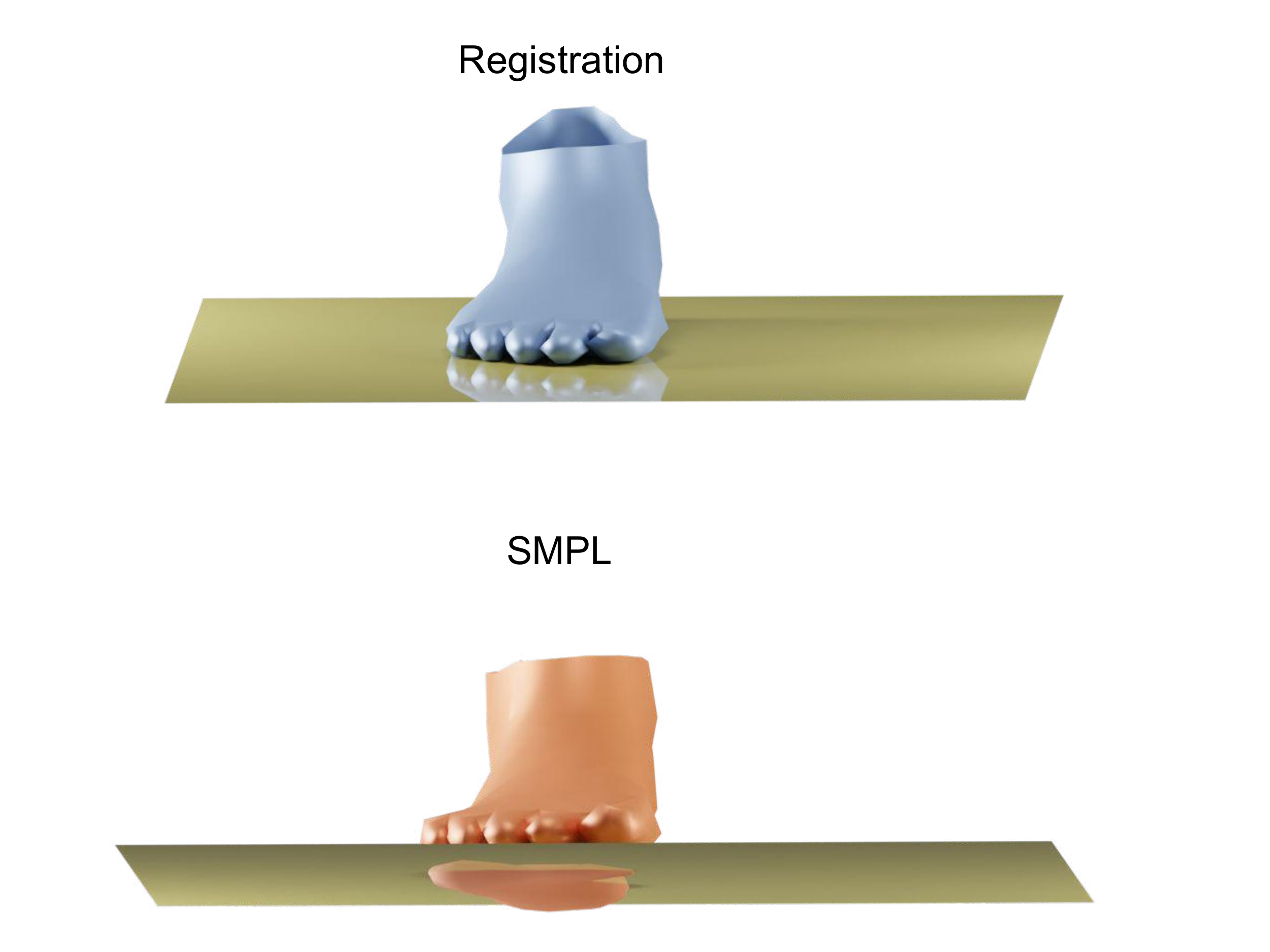}
 \caption{SMPL ground penetration.}
 \label{fig:contact_smpl}
\end{subfigure}
\caption{\textbf{Body part models failure cases.} Left: Existing body part models such as the \flame~\cite{FLAME:SiggraphAsia2017} head model and the \mano~\cite{MANO:SIGGRAPHASIA:2017} hand model fail to capture the corresponding body part's shape through the full range of motion. Fitting FLAME to a subject looking left  results in significant error in the neck region. Similarly, fitting  \mano  to hands with a bent wrist, results in significant error at the wrist region. Right: The foot of SMPL~\cite{SMPL:2015} fails to model  deformations due to ground contact, hence penetrating the ground. Additionally, it has a limited number of joints to model the toes articulation.}
\end{figure}

Hand~\cite{MANO:SIGGRAPHASIA:2017,Moon_2020_ECCV_DeepHandMesh,Smith2020,xu2020ghum}, head~\cite{cao2014facewarehouse,FLAME:SiggraphAsia2017,xu2020ghum} and body \cite{SMPL:2015,STAR:ECCV:2020} models are typically built independently. 
Heads and hands are captured with a 3D scanner in which a subject remains static, while the face and hands are articulated. This data is unnatural as it does not capture how the body parts move together with the body.
As a consequence, the construction of head/hand models implicitly assumes a static body, and use a simple kinematic tree that fails to model the head/hand full degrees of freedom. 
For example, in Fig.~\ref{fig:flame_mano_problems} we fit the \flame head model ~\cite{FLAME:SiggraphAsia2017} to a pose where the subject is looking right and find that \flame exhibits a significant error in the neck region. 
Similarly, we fit the \mano~\cite{MANO:SIGGRAPHASIA:2017} hand model to a hand pose where the the wrist is fully bent downwards. 
\mano  fails to capture the wrist deformation that results from the bent wrist. 
This is a systematic limitation of existing head/hand models, which can not be addressed by simply training on more data.

Another significant limitation of existing body-part models is the lack of an articulated foot model in the literature.
This is surprising given the many applications of a 3D foot model in the design, sale, and animation of footwear.
Feet are also critical for human locomotion.
Any biomechanical or physics-based model must have realistic feet to be faithful. 
The feet on existing full body models like SMPL are overly simplistic, have limited articulation, and do not deform with contact as shown in Fig.~\ref{fig:contact_smpl}.

In contrast to the existing approaches, we propose to jointly train the full human body and body part models together.
We first train a new full-body model called \modelname, with articulated hands and an expressive head using a federated dataset of body, hand, head and foot scans. 
This joint learning captures the full range of motion of the body parts along with the associated deformation.
Then, given the learned deformations, we separate the body model into body part models. 
To enable separating \modelname into compact individual body parts we learn a sparse factorization of the pose-corrective blend shape function as shown in the teaser Fig.~\ref{fig:teaser}. 
The factored representation of \modelname enables separating \modelname into an entire suite of models: \modelnameHead, \modelnameHand and \modelnameFoot. 
A body part model is separated by considering all the joints that influence the set of vertices defined by the body part template mesh. 
We show that the learned kinematic tree structure for the head/hand contains significantly more joints than commonly used by head/hand models. 
In contrast to the existing body part models that are learned in isolation of the body, our training algorithm unifies many disparate prior efforts and results in a suite of models that can capture the full range of motion of the head, hands, and feet.

\modelname goes beyond  existing statistical body models to include a novel foot model.
To do so, we extend the standard kinematic tree for the foot to allow more degrees of freedom.
To train the model, we capture foot scans using 
a custom 4D foot scanner (see Sup.~Mat.), where the foot is visible from all views, including the sole of the foot which is imaged through a glass plate.
This uniquely allows us to capture how the foot is deformed by contact with the ground.
We then model this deformation as a function of body pose and contact.

We train \modelname on $1.2$ million hand, head, foot, and body scans, which is an order of magnitude more data than the largest training dataset reported in the literature ($60$K \ghum ~\cite{xu2020ghum}). 
The training data contains extreme body shapes such as anorexia patients,  body builders, $14$K registrations from the CAESAR~\cite{CAESAR} and SizeUSA~\cite{SizeUSA} datasets and $7$K feet registrations from the ANSUR II dataset~\cite{ansur2}. 
All subjects gave informed written consent for participation and the use of their data in statistical models.
Capture protocols were reviewed by the local university ethics board.

We quantitatively compare \modelname and the individual body-part models to existing models including SMPL-X, \ghum, \mano, and \flame.  We find that \modelname is more expressive, is more accurate, and generalizes better.
In summary our main contributions are: 
(1) A unified framework for learning both expressive body models and a suite of high-fidelity body part models. 
(2) A novel 3D articulated foot model that captures compression due to contact.
(3) \modelname, a sparse expressive and compact body model that generalizes better than existing expressive human body models.
(4) An entire suite of body part models for the head, hand and feet, where the model kinematic tree and pose deformation are learned instead of being artist defined.
(5) The Tensorflow and a PyTorch implementations of all the models are publicly available for research purposes. 
\section{Related Work}
\label{sec:related_work}

\textbf{Body Models:} 
SCAPE \cite{Anguelov05} is the first 3D model to factor body shape into separate pose and a shape spaces.
SCAPE is based on triangle deformations and is not compatible with existing graphics pipelines.
In contrast,
\smpl \cite{SMPL:2015} is the first learned statistical body model compatible with game engines
SMPL is a vertex-based model with linear blend skinning (LBS) and learned pose and shape corrective blendshapes. 
A key drawback of SMPL is that it relates the pose corrective blendshapes to the elements of the part rotations matrices of all the model joints in the kinematic tree. Consequently, it learns spurious long-range  correlations in the training data. 
STAR \cite{STAR:ECCV:2020} addresses many of the drawback of SMPL by using a compact representation of the kinematic tree based on quaternions and learning sparse pose corrective blendshapes where each joint strictly influences a sparse set of the model vertices. 
The pose corrective blendshape formulation in \modelname is based on STAR.
Also related to our work, the \emph{Stitched Puppet} \cite{zuffi2015stitched} is a part-based model of the human body. The body is segmented into $16$ independent parts with learned pose and shape corrective blendshapes. A pairwise stitching function fuses the parts, but leaves visible discontinuities.
While \modelname is also part-based model, we start with a unified model and learn its segmentation into parts during training from a federated training dataset.

\textbf{Expressive Body Models:} The most related to \modelname are expressive body models such as Frank~\cite{joo2018total}, \smplx~\cite{SMPL-X:2019}, and GHUM \& GHUML \cite{xu2020ghum,zanfir}. 
Frank \cite{joo2018total} merges the body of \smpl~\cite{SMPL:2015} with the FaceWarehouse~\cite{cao2014facewarehouse} face model and an artist-defined hand rig. Due to the fusion of different models learned in isolation, Frank looks unrealistic. 
\smplx~\cite{SMPL-X:2019} learns an expressive body model and fuses the MANO hand model~\cite{MANO:SIGGRAPHASIA:2017} pose blendshapes and the FLAME head model  ~\cite{FLAME:SiggraphAsia2017} expression space. 
However, since MANO and FLAME are learned in isolation of the body, they do not capture the full degrees of freedom of the head and hands. Thus, fusing the parameters results in artifacts at the boundaries. In contrast to the construction of Frank and \smplx, for \modelname, we start with a coherent full body model, trained on a federated dataset of body, hand, head and feet scans, then separate the model into individual body parts. 
Xu et al.~\cite{xu2020ghum} propose GHUM \& GHUML, which are trained on a federated dataset of $60K$ head, hand and body scans and use a fully connected neural network architecture to predict the pose deformation. The \ghum model can not be separated into body parts as a result of the dense fully connected formulation that relates all the vertices to all the joints in the model kinematic tree. In contrast, the \modelname factorized representation of the pose space deformations enables seamless separation of the body into head/hand and foot models.

\textbf{Head Models:} There are many models of 3D head shape \cite{blanz1999morphable,Booth2018,paysan20093d}, shape and expression \cite{Amberg2008,Brunton2014,cao2014facewarehouse,li2020learning,COMA:ECCV18,yang2020facescape,Vlasic2005} or shape, pose and expression \cite{FLAME:SiggraphAsia2017}. We focus here on models with a full head template, including a neck. 
The FLAME head model \cite{FLAME:SiggraphAsia2017},  like \smpl, uses a dense pose corrective blendshape formulation that relates all vertices to all joints. Xu et al.~\cite{xu2020ghum} also propose \ghumHead, where the template is based on the \ghum head with a retrained pose dependant corrector network (PSD). Both \ghumHead and \flame are trained in isolation of the body and do not have sufficient joints to model the full head degrees of freedom. In contrast to the previous methods, \modelnameHead is trained jointly with the body on a federated dataset of head and body meshes, which is critical to model the head full range of motion.
It also has  more joints than \ghumHead or \flame, which we show is crucial to model the head full range of motion.

\textbf{Hand Models:} 
\mano \cite{MANO:SIGGRAPHASIA:2017} is widely use and is based on the \smpl formulation where the pose corrective blendshapes deformations are regularised to be local. The kinematic tree of \mano is based on spherical joints allowing redundant degrees of freedom for the fingers. Xu et al.~\cite{xu2020ghum} introduce the \ghumHand model where they separate the hands from the template mesh of \ghum and train a hand-specific pose-dependant corrector network (PSD). Both \mano and \ghumHand are trained in isolation of the body and result in implausible deformation around the wrist area. \modelnameHand is trained jointly with the body and has a wrist joint which is critical 
to model the hands full range of motion. 

\textbf{Foot Models:} Statistical shape models of the feet are less studied than those of the body, head, and hands. Conard et al.~\cite{conrad2019statistical} propose a statistical shape model of the human foot, which is a PCA space learned from static foot scans. 
However, the human feet deform with motion and models learned from static scans can not capture the complexity of 3D foot deformations.  
To address the limitations of static scans,
Boppana et al.~\cite{boppana2021dynamic} propose the DynaMo system to capture scans of the feet in motion and learn a PCA-based model from the scans.  However, the DynaMo setup fails to capture the sole of the foot in motion. 
In contrast, to all prior work, \modelnameFoot contains a kinematic tree, a pose deformation space, and a PCA shape space. We use a specialized 4D foot scanner, where the entire human foot is visible and accurately reconstructed, including the toes and the sole. 
Furthermore, we go beyond previous work to model the foot deformations resulting from ground contact, which was not possible before.
\section{Model}
\label{sec:model}

We describe the formulation of \modelname in  Section~\ref{sec:supr}, followed by how we separate \modelname into body parts models in Section~\ref{sec:body_part}. Since existing body corrective deformation formulations fail to model foot deformations due to ground contact, we discuss a novel foot deformation network in Section~\ref{sec:foot}.

\begin{figure}[tbh]
	\centering
	\begin{subfigure}[b]{0.30\linewidth}
 \includegraphics[width=\textwidth]{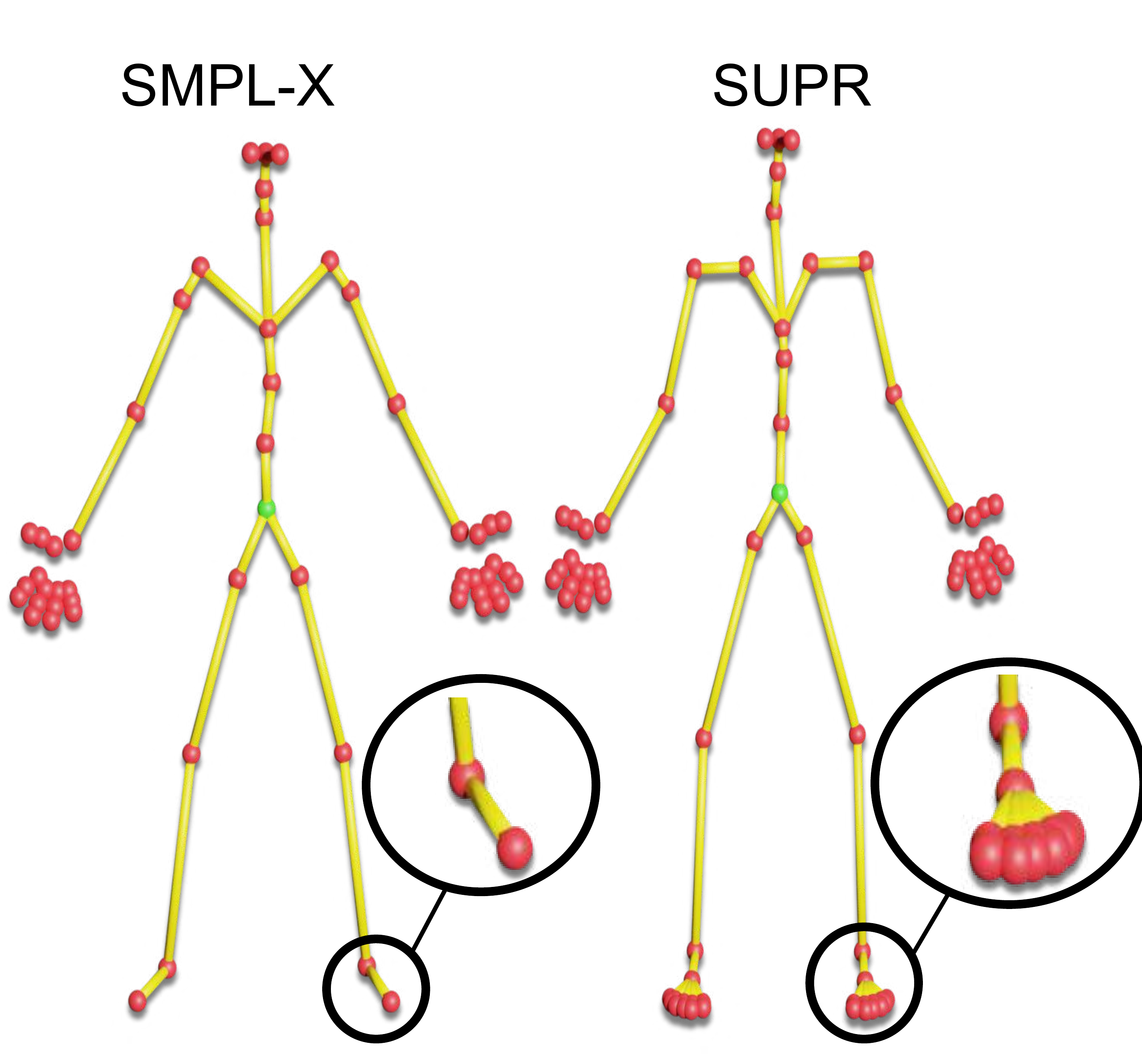}
 \caption{Body}
 \label{fig:supr_kintree}
\end{subfigure}
	\begin{subfigure}[b]{0.30\linewidth}
 \includegraphics[width=\linewidth]{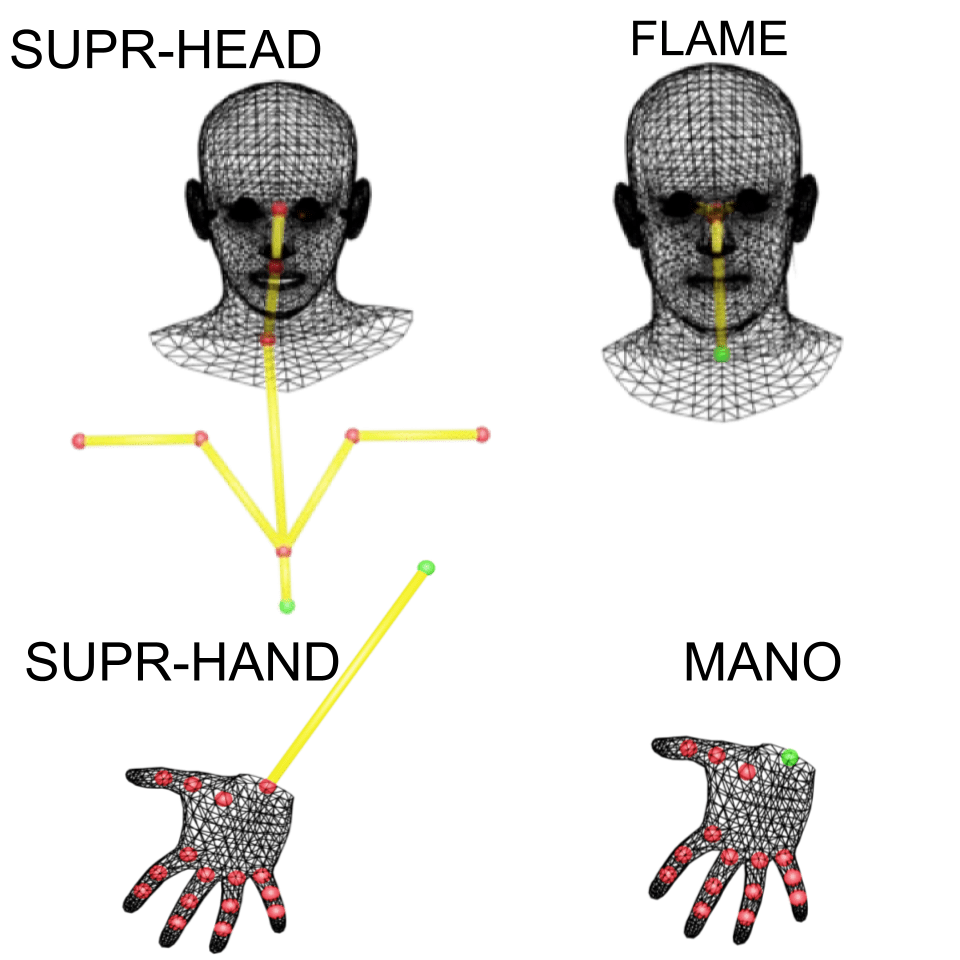}
 \caption{Head/Hand}
  \label{fig:supr_separated}
\end{subfigure}
\begin{subfigure}[b]{0.30\linewidth}
 \includegraphics[width=\linewidth]{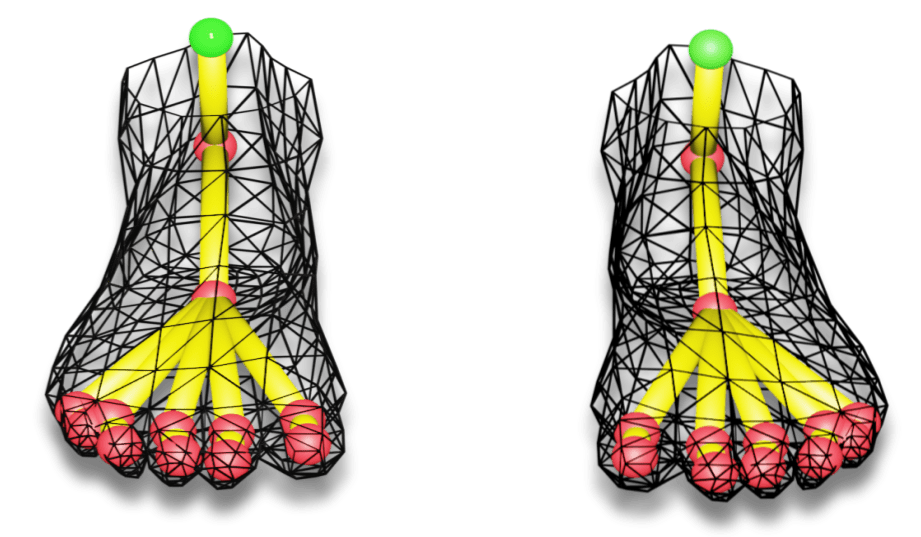}
 \caption{\modelnameFoot}
\end{subfigure}
\caption{The kinematic tree of \modelname and the separated body part models. }
\label{fig:kinematic_tree}
\end{figure}

\subsection{SUPR}
\label{sec:supr}
\modelname is a vertex-based 3D model with linear blend skinning (LBS) and learned blend shapes. 
The blend shapes are decomposed into 3 types: \emph{Shape Blend Shapes} to capture the subject identity, \emph{Pose-Corrective Blend Shapes} to correct for the widely-known LBS artifacts, and \emph{Expression  Blend Shapes} to model facial expressions. The \modelname mesh topology and kinematic tree are based on the \smplx topology. The template mesh contains $N=10,475$  vertices and $K = 75$ joints. The \modelname kinematic tree is shown in Figure \ref{fig:kinematic_tree}. 
In contrast to existing body models, the \modelname kinematic tree contains significantly more joints in the foot, ankle and toes as shown in Fig.~\ref{fig:supr_kintree}. 
Following the notation of \smpl,
\modelname  is defined by a function $M(\vec{\theta},\vec{\beta},\vec{\psi})$, where $\vec{\theta} \in \mathbb{R}^{75 \times 3}$ are the pose parameters corresponding to the individual bone rotations, $\vec{\beta} \in \mathbb{R}^{300}$ are the shape parameters corresponding to the subject identity, $\vec{\psi} \in \mathbb{R}^{100}$ are the expression parameters controlling facial expressions.  Formally, \modelname is defined as
\begin{equation}
	M(\vec{\theta},\vec{\beta},\vec{\psi})=W(T_p(\vec{\theta},\vec{\beta},\vec{\psi}),J(\vec{\beta}),\vec{\theta};\mathcal{W}),
	\label{Eq:supr}
\end{equation}
where the 3D body,  $T_p(\vec{\theta},\vec{\beta},\vec{\psi})$,  is  transformed around the joints $J$ by the linear-blend-skinning function $W(.)$, parameterized by the skinning weights $\mathcal{W} \in \mathbb{R}^{10475 \times 75}$. The cumulative corrective blend shapes term is defined as
\begin{equation}
	T_p(\vec{\theta},\vec{\beta},\vec{\psi}) = \overline{T} + B_S(\vec{\beta};\mathcal{S}) +  B_P(\vec{\theta};\mathcal{P}) +  B_E(\vec{\psi};\mathcal{E}),
	\label{Eq:STAR}
\end{equation}
where $\overline{T} \in \mathbb{R}^{10475\times3}$  is the template of the mean body shape, which is  deformed by: $B_S(\vec{\beta};\mathcal{S})$, the shape blend shape function capturing a PCA space of body shapes; $B_P(\vec{\theta};\mathcal{P}) $, the pose-corrective blend shapes that address the LBS artifacts; and $B_E(\vec{\psi};\mathcal{E})$,  a PCA space of facial expressions. \par  

\subsubsection{Sparse Pose Blend Shapes}
In order to separate \modelname into body parts, each joint should strictly influence a subset of the template vertices $\overline{T}$. To this end, we base the pose-corrective blend shapes $B_p(.)$ in Eq.~\ref{Eq:STAR} on the STAR model \cite{STAR:ECCV:2020}. The pose-corrective blend shape function is factored into per-joint pose corrective blend shape functions
\begin{equation}
	\label{eq:supr blends_blends}
	B_P({\vec{q}},\mathbf{K},\mathbf{A}) = \sum_{j=1}^{K-1} {B}^{j}_P({\vec{q}_{ne(j)}};{\mathbf{K}_j};{A_j}),
\end{equation}
where the  pose-corrective blend shapes are sum of $K-1$ sparse spatially-local pose-corrective blend-shape functions. Each joint-based corrective blend shape ${B}^{j}_P(.)$, predicts corrective offsets for a sparse set of the model vertices, defined by the learned joint activation weights  $A_j \in \mathbb{R}^{10475}$. 
Each $A_j$ is a sparse vector defining the sparse set of vertices influenced by the $j^{\mathrm{th}}$ joint blend shape $B^{j}_p(.)$. The joint corrective blend shape function is conditioned on the normalized unit quaternions  $\vec{q}_{ne(j)}$ of the $j^{\mathrm{th}}$ joint's direct neighbouring joints' pose parameters. We note that the  \modelname pose blend-shape formulation in Eq.~\ref{eq:supr blends_blends} is not conditioned on body shape, unlike STAR, since the additional body-shape blend shape is not sparse and, hence, can not be factorized into body parts. Since the skinning weights in Eq.~\ref{Eq:supr} and the pose-corrective blend-shape formulation in Eq.~\ref{eq:supr blends_blends} are sparse, each vertex in the model is related to a small subset of the model joints. This sparse formulation of the pose space is key to separating the model into compact body part models.

\subsection{Body Part Models}
\label{sec:body_part}
In traditional body part models like \flame and \mano, the kinematic tree is designed by an artist and the models are learned in isolation of the body. 
In contrast, here the pose-corrective blend shapes of the hand (\modelnameHand), head (\modelnameHead) and foot (\modelnameFoot) models are trained jointly with the body on a federated dataset. 
The kinematic tree of each part model is inferred from \modelname rather than being artist defined. 
To separate a body part, we first define the subset of mesh vertices of the body part $\overline{T}_{bp}$
from the \modelname template $\overline{T}_{bp} \in \overline{T}$. Since the learned \modelname skinning weights and pose-corrective blend shapes are strictly sparse, any subset of the model vertices $\overline{T}_{bp}$ is strictly influenced by a subset of the model joints. More formally, a joint $\vec{j}$ is deemed to influence a body part defined by the template $\overline{T}_{bp}$ if:
\begin{equation}
    \mathbb{I}\big(T_{bp},\vec{j}\big) = \begin{cases}
  1 & \small\text{if $\sum \mathcal{W}\big(\overline{T}_{bp},\vec{j}\big) \neq 0 $ or $\sum A_j\big(\overline{T}_{bp}\big) \neq 0$}\\ 
  0 & \text{othewise},
\end{cases} 
\end{equation}

where $\mathbb{I}(.,.)$ is an indicator function, $\mathcal{W}\big(\overline{T}_{bp} ,\vec{j}\big)$ is a subset of the \modelname learned skinning weights matrix, where the rows are defined by the vertices of $\overline{T}_{bp}$, the columns correspond to the $j^{\mathrm{th}}$ joint, $\vec{j}$, $A_j(\overline{T}_{bp})$ corresponds to the learned activation for the $j^{\mathrm{th}}$ joint and the rows defined by vertices $\overline{T}_{bp}$. The indicator function $\mathbb{I}$ returns $1$ if a joint $\vec{j}$ has non-zero skinning weights or a non-zero activation for the vertices defined by $\overline{T}_{bp}$. Therefore the set of joints $J_{bp}$ that influences the template $\overline{T}_{bp}$ is defined by: 
\begin{equation}
    J_{bp} = \Big\{ \mathbb{I}(\overline{T}_{bp},j) = 1 \hspace{0.2cm} \forall \hspace{0.2cm}
 j \in \{1,\dots,K \} \Big\}.
 \label{eq:rig}
\end{equation}
The kinematic tree  defined for the body part models in Eq.~\ref{eq:rig} is implicitly defined by the learned skinning weights $\mathcal{W}$  and the per joint activation weights $A_j$. The resulting kinematic tree of the separated models is shown in Fig.~\ref{fig:supr_separated}. Surprisingly, the head is influenced by substantially more joints than in the artist-designed kinematic tree used in \flame. Similarly, \modelnameHand has an additional wrist joint compared to \mano. We note here that the additional joints in \modelnameHead and \modelnameHand are outside the head/hand mesh. The additional joints for the head and the hand are beyond the scanning volume of a body part head/hand scanner. 
This means that it is not possible to learn the influence of the shoulder and spine joints on the neck from head scans alone. 

The skinning weights for a separated body are defined by $\mathcal{W}_{bp} = \mathcal{W}\big(\overline{T}_{bp},J_{bp}\big)$,
where $\mathcal{W}\big(\overline{T}_{bp},J_{bp}\big)$ is the subset of the \modelname skinning weights defined by the rows corresponding to the vertices of $\overline{T}_{bp}$ and the columns defined by $J_{bp}$. Similarly, the pose corrective blendshapes are defined by $
 B_{bp}=B_{p}\big(\overline{T}_{bp},J_{bp}\big)$ 
where $B_{p}\big(\overline{T}_{bp},J_{bp}\big)$ corresponds to a subset of \modelname pose blend shapes defined by the vertices of $\overline{T}_{bp}$ and the quaternion features for the set of joints $J_{bp}$. The skinning weights $\mathcal{W}_{bp}$ and blendshapes $B_{bp}$ are based on the \modelname learned blend shapes and skinning weights, which are trained on a federated dataset that explores each body part's full range of motion relative to the body. We additionally train a joint regressor  $\mathcal{J}_{bp}$, to regress the joints $\mathcal{J}_\text{bp}: \overline{T}_{_{bp}} \rightarrow J_{bp}$. We learn a local body part shape space $B_S(\vec{\beta}_{bp};\mathcal{S}_{bp})$, where $\mathcal{S}_{bp}$ is the body part PCA shape components. For the head, we use the \modelname learned expression space $B_E(\psi;\mathcal{E})$. 

\subsection{Foot deformation Network} 
\label{sec:foot}
The linear pose-corrective blend shapes in Eq.~\ref{Eq:STAR} and Eq.~\ref{eq:supr blends_blends} relate the body deformations to the body pose only. However, the human foot deforms as a function of pose, shape and ground contact. 
To model this, we add a foot deformation network.

The foot body part model, separated from \modelname, is defined by the pose parameters $\vec{\theta}_{bp} \in \vec{\theta}$, corresponding to the ankle and toe pose parameters in addition to $\vec{\beta}_{bp}$, the PCA coefficients of the  local foot shape space. We extend the pose blend shapes in Eq.~\ref{Eq:STAR} to include a deep corrective deformation term for the foot vertices defined by $\overline{T}_{\mathit{foot}} \in \overline{T}$. With a slight abuse of notation, we will refer to the deformation function $T_p(\vec{\theta},\vec{\beta},\vec{\psi})$ in Eq.~\ref{Eq:STAR} as $T_p$ for simplicity. The foot deformation function is defined by:
\begin{equation}
  T_p'(\vec{\theta},\vec{\beta},\vec{c}) = T_p +  \vec{m} \circ B_F(\vec{\theta}_{\mathit{foot}},\vec{\beta}_{\mathit{foot}},\vec{c};\mathcal{F}),
\end{equation}
where $\vec{m} \in \{0,1\}^{10475}$ is a binary vector with ones corresponding to the foot vertices and $0$  elsewhere. $B_F(.)$ is a multilayer perceptron-based deformation function parameterized by $\mathcal{F}$, conditioned on the foot pose parameters $\vec{\theta}_{\mathit{foot}}$, foot shape parameters $\vec{\beta}_{\mathit{foot}}$ and foot contact state $\vec{c}$. The foot contact state variable is a binary vector $\vec{c} \in \{0,1\}^{266}$ defining the contact state of each vertex in the foot template mesh, a vertex is represented by a $1$ if it is in contact with the ground, and 0 otherwise. The Hadamard product between $\vec{m}$ and $B_F(.)$ ensures the network $B_F(.)$ strictly predicts deformations for the foot vertices only.

\textbf{Implementation details.} The foot contact deformation network is based on an encoder-decoder architecture.
The input feature,  $\vec{f} \in \mathbb{R}^{320}$, to the encoder is a concatenated feature of the foot pose, shape and contact vector. The foot pose is represented with a normalised unit quaternion representation, shape is encoded with the first two PCA coefficients of the local foot shape space. The input feature $\vec{f}$ is encoded into a latent vector $\vec{z} \in \mathbb{R}^{16}$ using fully connected layers with a leaky LReLU as an activation function with a slope of $0.1$ for negative values. The latent embedding $\vec{z}$ is decoded to predict deformations for each vertex using fully connected layers with LReLU activation. The full architecture is described in detail in \supmat 
We train male, female and a gender-neutral versions of \modelname and the separated body part models. Training details are discussed in \supmat
\begin{figure}
\centering
    
     \begin{subfigure}{\linewidth}
         \begin{subfigure}[b]{\linewidth}
             \includegraphics[width=0.9\linewidth]{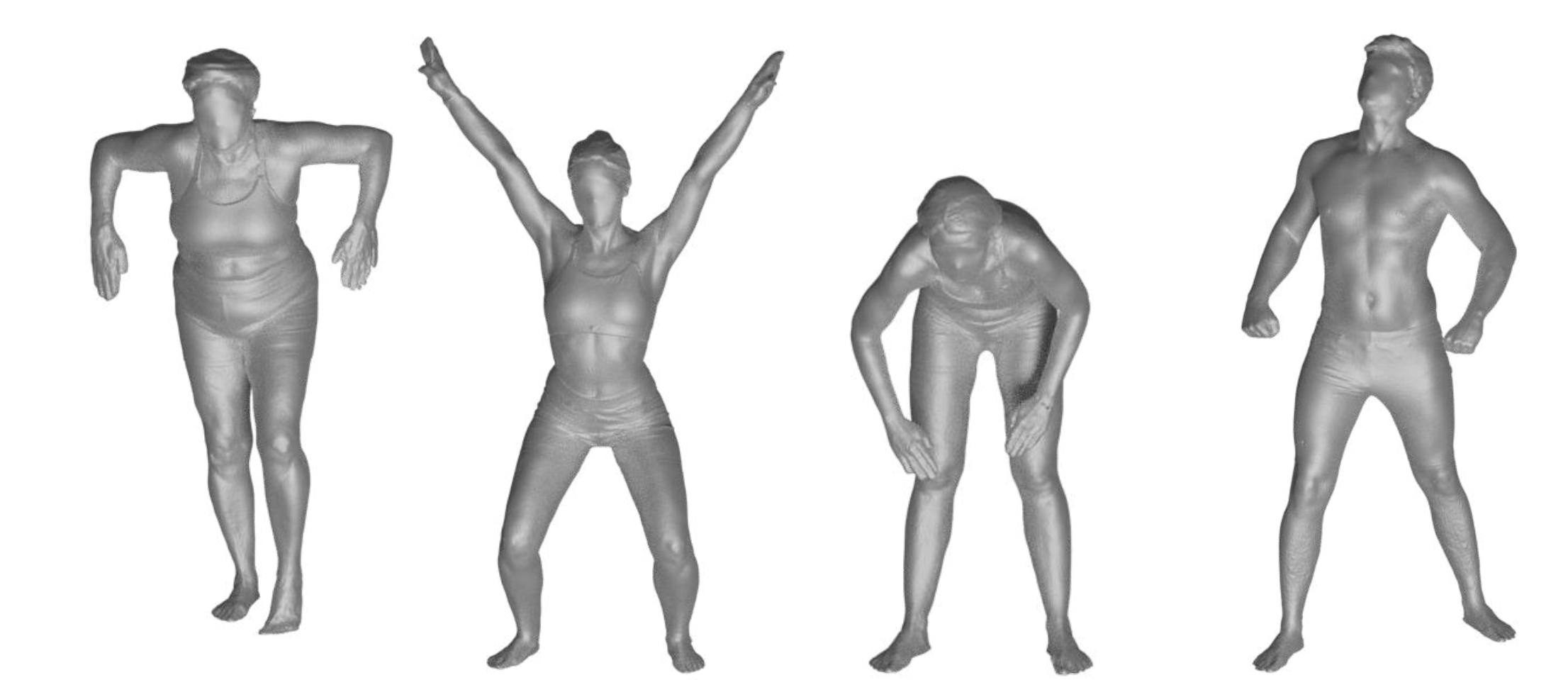}
             \caption{3DBodyTex scans.}
             \label{fig:raw_scans}
         \end{subfigure}
         \begin{subfigure}[b]{0.85\linewidth}
             \includegraphics[width=\linewidth]{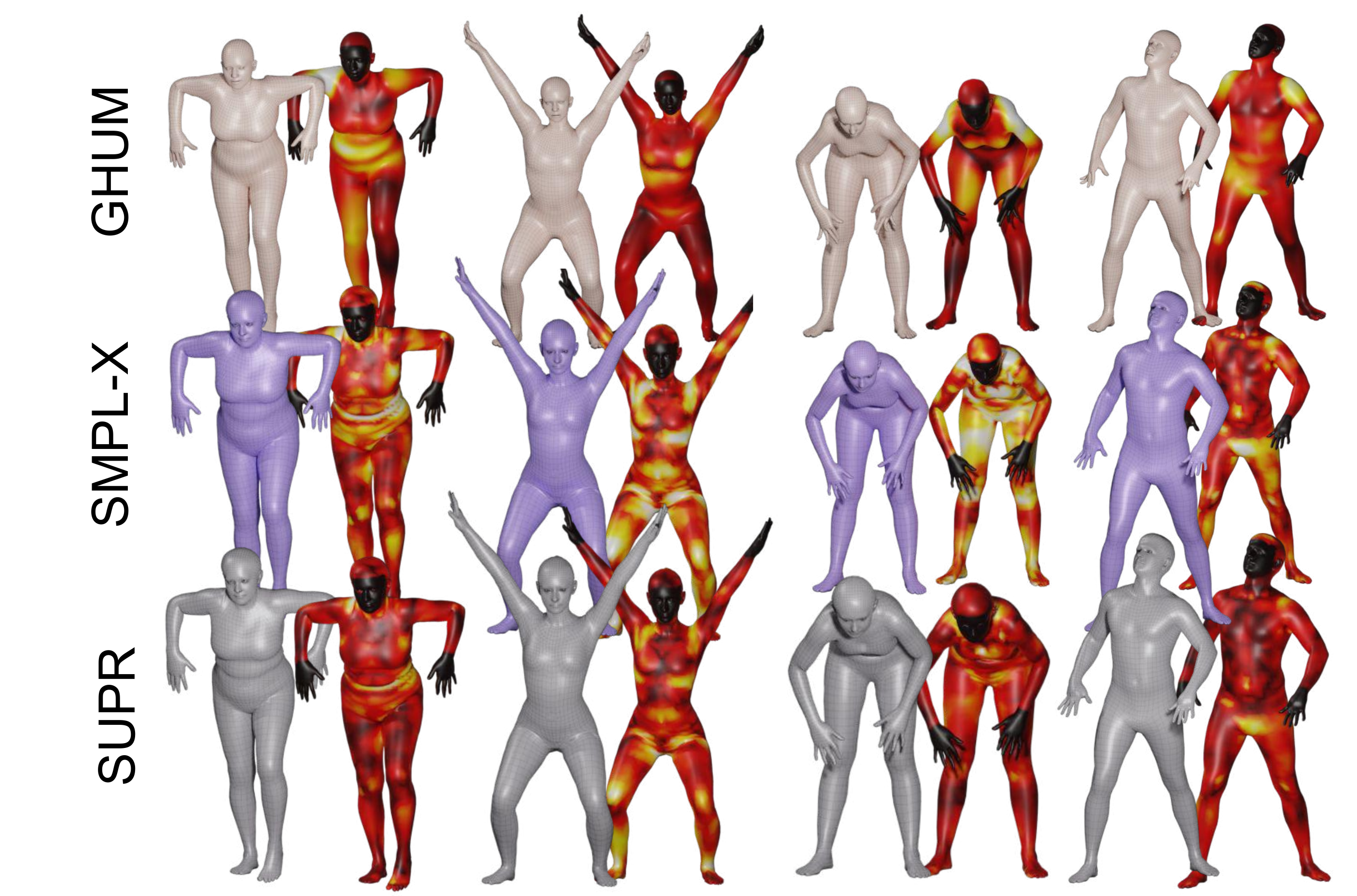}
             \caption{Model Fits}
             \label{fig:body_fits}
         \end{subfigure}
                   \includegraphics[width=1.2cm,height=2.5in]{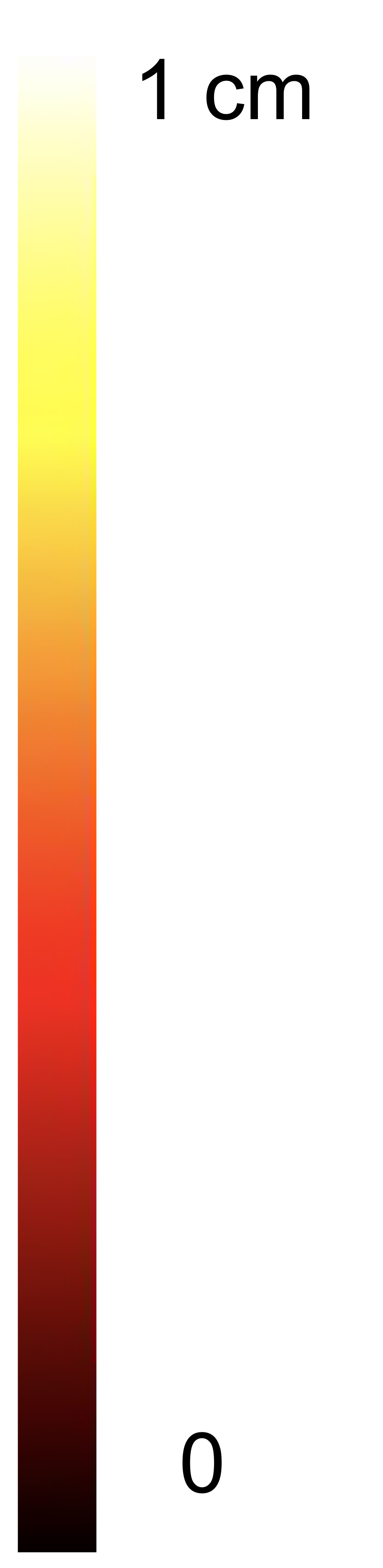}
	\end{subfigure}
	
     \begin{subfigure}{\linewidth}
    \begin{subfigure}[b]{0.5\linewidth}
    \includegraphics[width=0.8\linewidth]{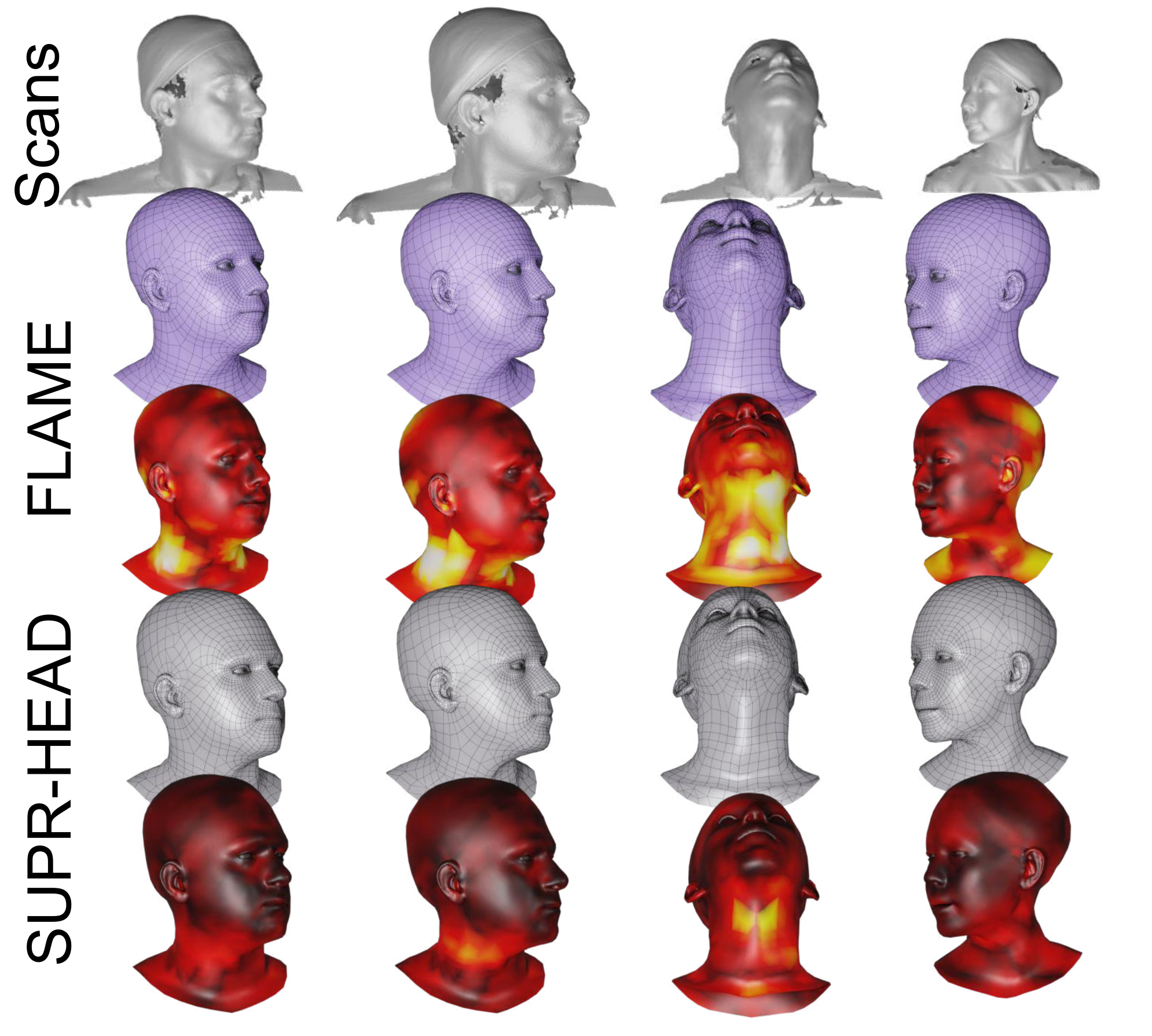}
    \caption{Head Evaluation}
    \label{fig:head_eval}
    \end{subfigure}
       \begin{subfigure}[b]{0.5\linewidth}
    \includegraphics[width=0.8\linewidth]{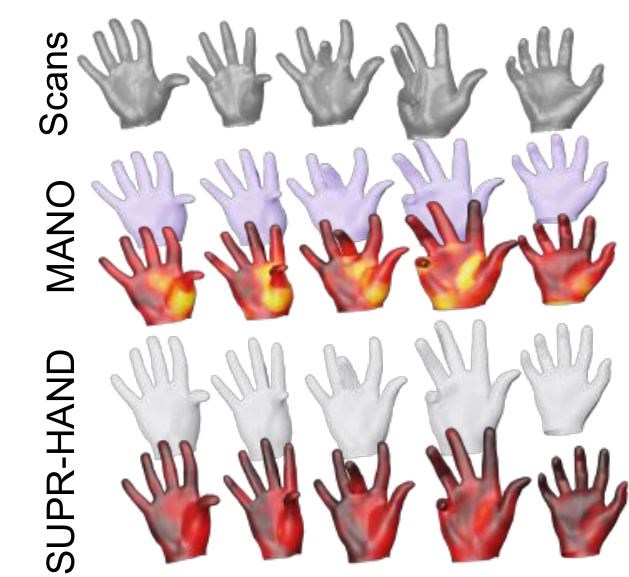}
    \caption{Hand Evaluation}
        \label{fig:hand_eval}
    \end{subfigure}
        \end{subfigure}
\caption{\textbf{Qualitative Evaluation:} We evaluate \modelname and the separated body part models against baselines. We use the 3DBodyTex dataset in Fig.~\ref{fig:raw_scans} to evaluate GHUM, SMPL-X and \modelname in Fig.~\ref{fig:body_fits} using $16$ shape components. We evaluate \modelnameHead against FLAME in Fig.~\ref{fig:head_eval} using $16$ shape components and \modelnameHand against MANO in Fig.~\ref{fig:hand_eval} using $8$ shape components.}
\end{figure}

\begin{figure*}[t]
     \centering
     \begin{subfigure}[b]{0.24\textwidth}
         \centering
         \includegraphics[width=1.15\textwidth]{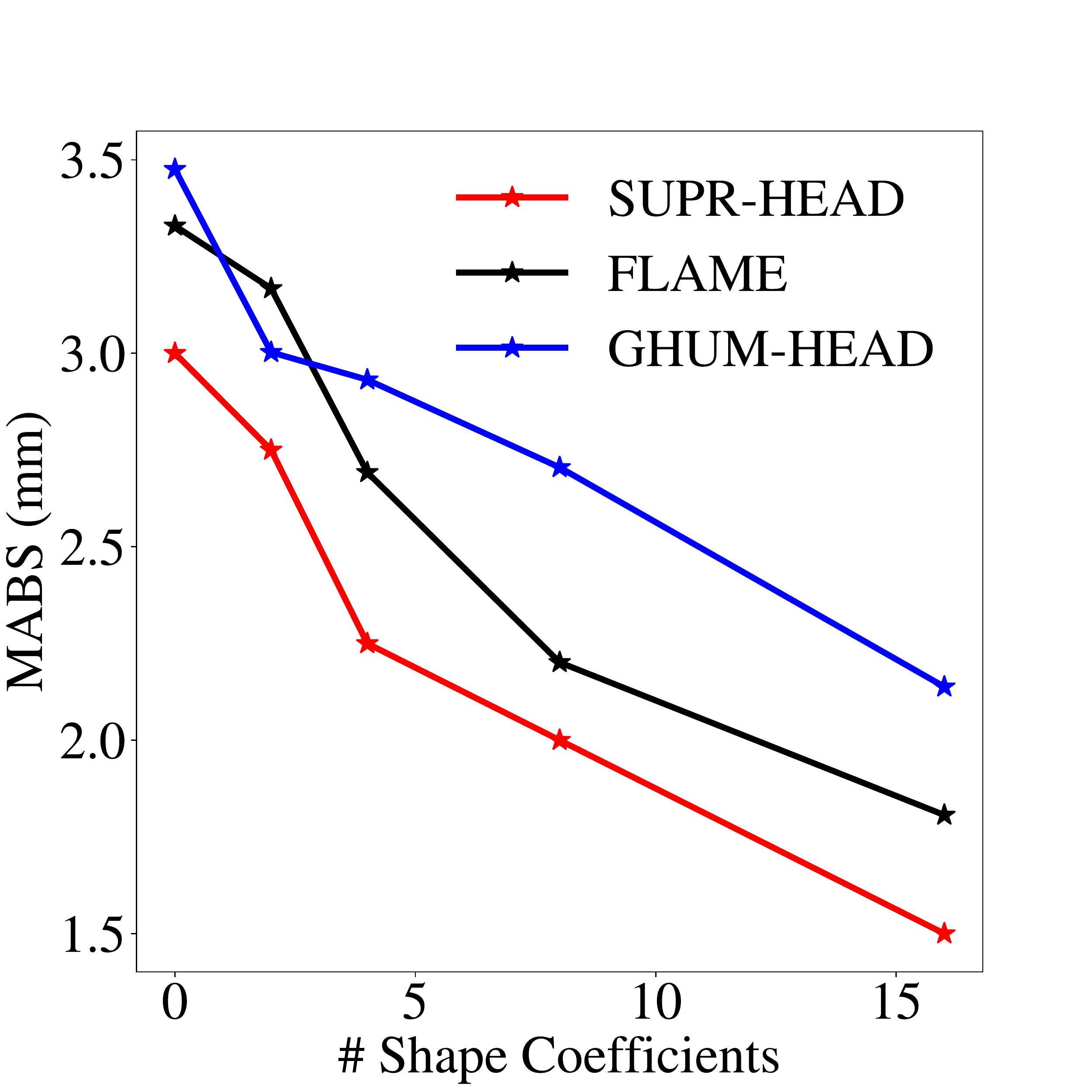}
         \caption{Head Evaluation}
         \label{fig:head_num}
     \end{subfigure}
     \begin{subfigure}[b]{0.24\textwidth}
         \centering
         \includegraphics[width=1.15\textwidth]{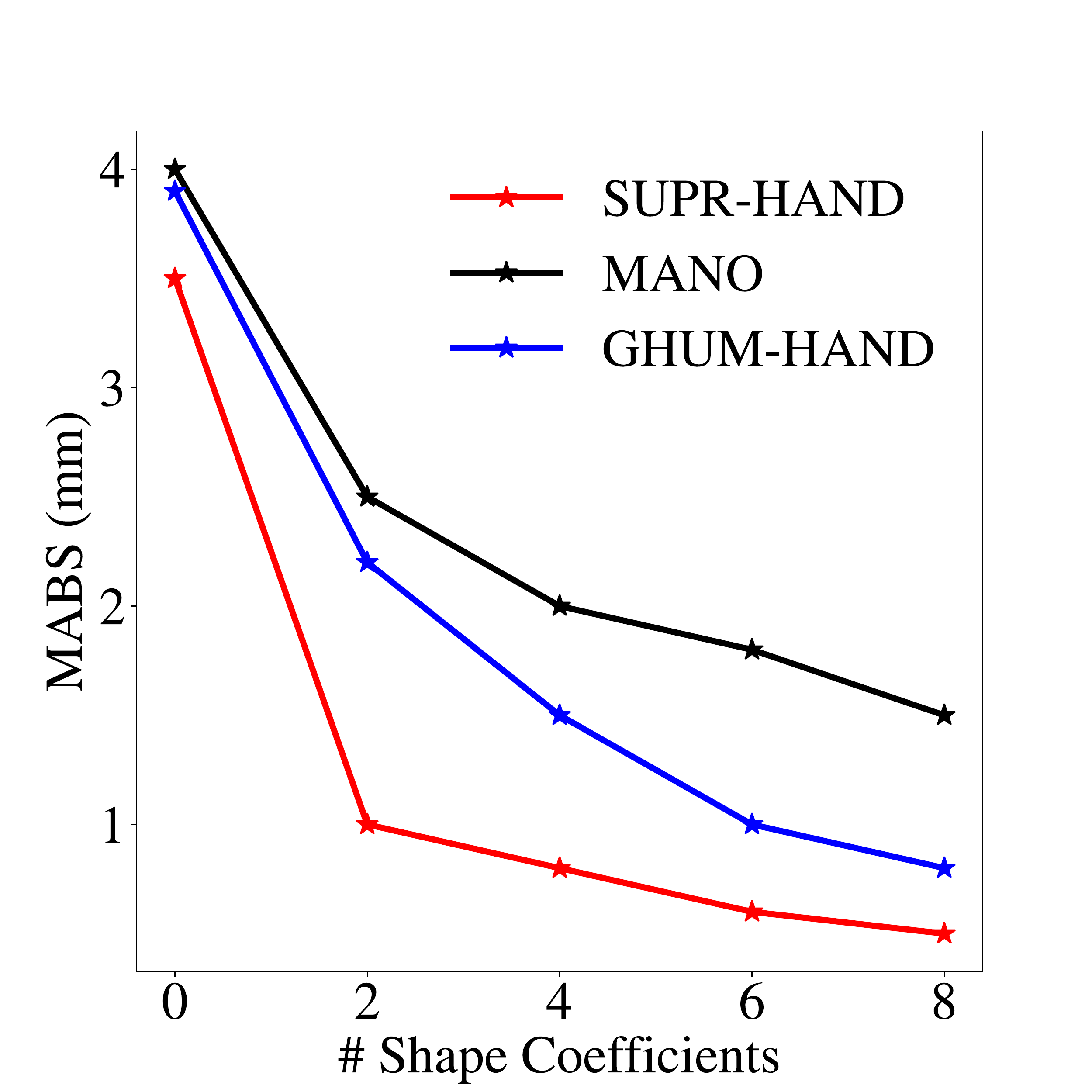}
         \caption{Hand Evaluation}
         \label{fig:hand_num}
     \end{subfigure}
     \begin{subfigure}[b]{0.24\textwidth}
         \centering
         \includegraphics[width=1.15\textwidth]{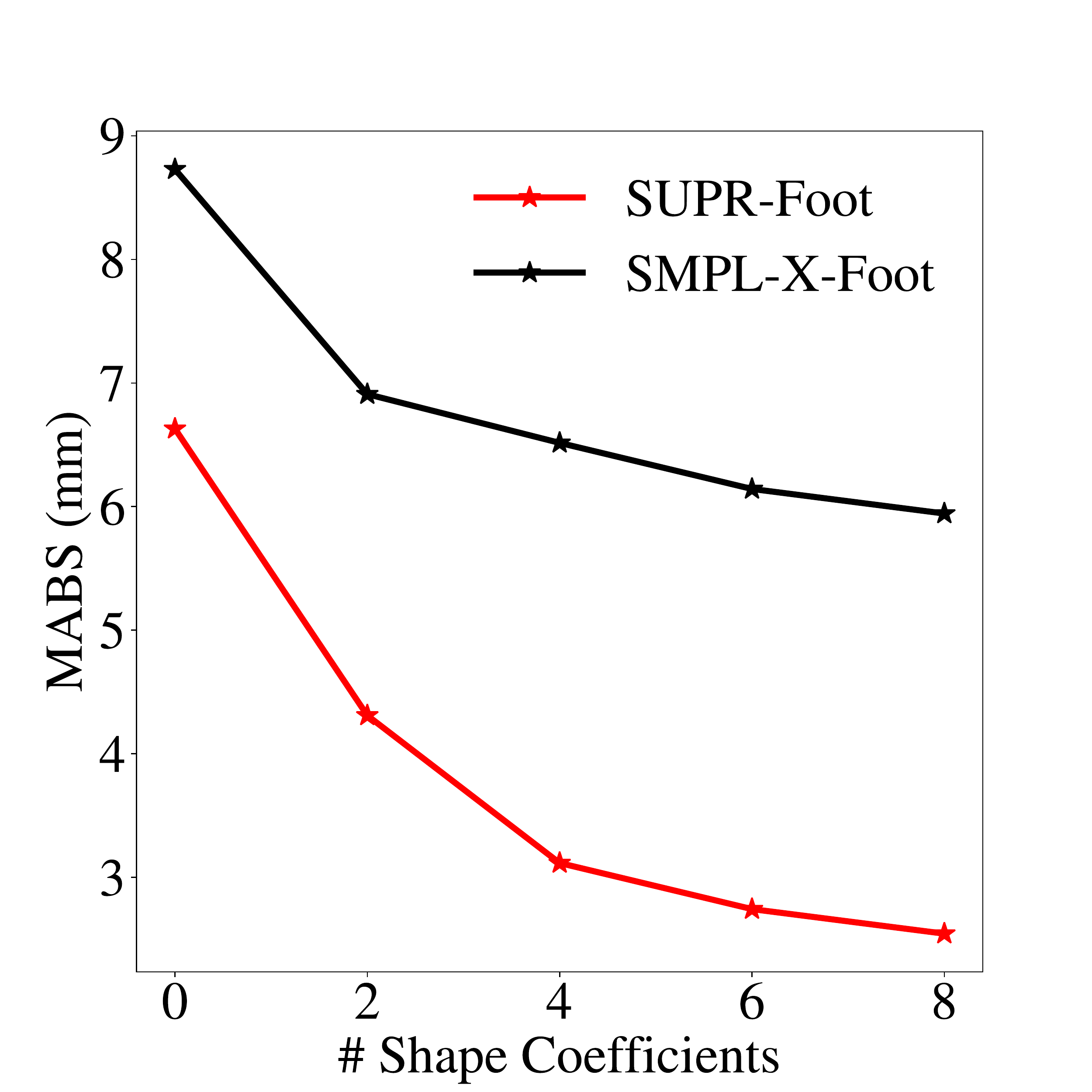}
         \caption{Foot Evaluation}
         \label{fig:foot_num}
     \end{subfigure}
     \begin{subfigure}[b]{0.24\textwidth}
         \centering
         \includegraphics[width=1.15\textwidth]{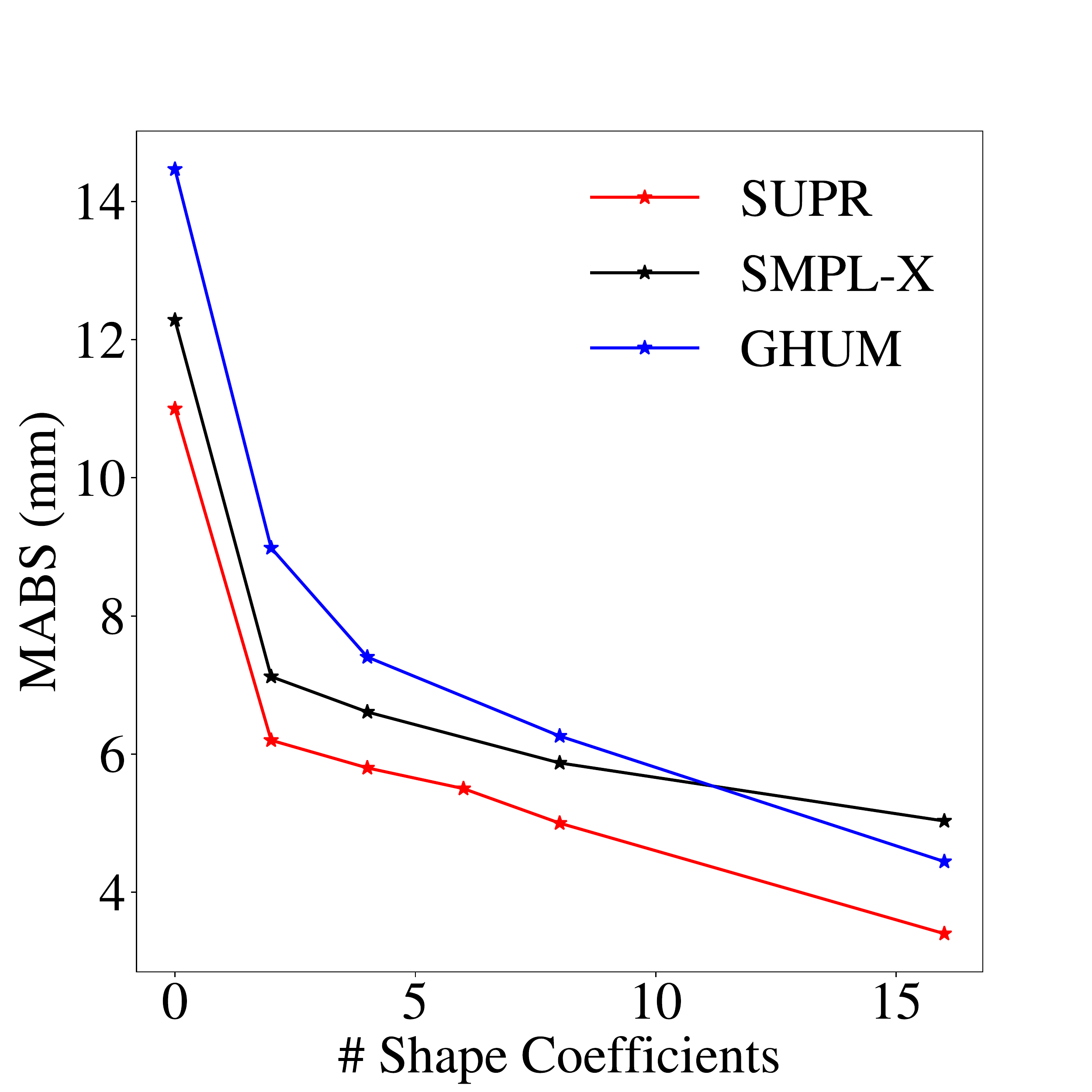}
         \caption{Body Evaluation}
         \label{fig:body_num}
     \end{subfigure}
        \caption{\textbf{Quantitative Evaluation:} Evaluating the generalization of the separated head, hand and foot model from \modelname against existing body part models: GHUM-HEAD and \flame for the head (Fig.~\ref{fig:head_num}), GHUM-HAND and \mano (Fig.~\ref{fig:hand_num}). We report the \emph{vertex-to-vertex} error ($mm$) as a function of the number of the shape coefficients used when fitting each model to the test set.}
        \label{fig:three graphs}
\end{figure*}
\section{Experiments}
\label{sec:experiments}
 Our goal is to evaluate the generalization of \modelname and the separated head, hand, and foot model to unseen test subjects. We first evaluate the full \modelname body model against existing state of the art expressive human body models \smplx and \ghum (Section~\ref{body_eval}),
then we evaluate the separated \modelnameHead model against existing head models \flame and \ghumHead (Section~\ref{head_eval}), and compare the hand model to \ghumHand and \mano (Section~\ref{hand_eval}). Finally, we evaluate the \modelnameFoot (Section~\ref{foot_eval}).

\subsection{Full-Body Evaluation}
\label{body_eval}
We use the publicly available 3DBodyTex dataset \cite{saint20183dbodytex}, which includes 100 male and 100 female subjects. 
We register the \ghum template and the \smplx template to all the scans; note \smplx and \modelname share the same mesh topology. 
We visually inspected all registered meshes for quality control.
Given registered meshes, we fit each model by minimizing the vertex-to-vertex loss (\emph{v2v}) between the model surface and the corresponding registration. 
The free optimization parameters for all models are the pose parameters $\vec{\theta}$ and the shape parameters $\vec{\beta}$.  
Note that, for fair comparison with \ghum, we only report errors for up to 16 shape components since this is the maximum in the \ghum release.  
\modelname includes 300 shape components that would reduce the errors significantly.

We follow the 3DBodyTex  evaluation protocol and exclude the face and the hands when reporting the mean absolute error (\emph{mabs}). We report the mean absolute error of each model on both male and female registrations. For the \ghum model, we use the PCA-based shape and expression space. We report the model generalization error in Fig.~\ref{fig:body_num} and show a qualitative sample of the model fits in Fig.~\ref{fig:body_fits}. \modelname uniformly exhibits a lower error than \smplx and \ghum.
\subsection{Head Evaluation}
\label{head_eval}
The head evaluation test set contains a total of $3$ male and $3$ female subjects, with sequences containing extreme facial expression, jaw movement and neck movement. As for the full body, we register the \ghumHead model and  the \flame template to the test scans, and use these registered meshes for evaluation.
For the \ghumHead model, we use the linear PCA expression and shape space. We evaluate all models using a standard \emph{v2v} objective, where the optimization free variables are the model pose, shape parameters, and expression parameters. We use $16$ expression parameters when fitting all models. For \ghumHead we exclude the internal head geometry (corresponding to a tongue-like structure) when reporting the $v2v$ error. Fig.~\ref{fig:head_num} shows the model generalization as a function of the number of shape components. We show a sample of the model fits in Fig.~\ref{fig:head_eval}. Both  \ghumHead and FLAME fail to capture head-to-neck rotations plausibly, despite each featuring a full head mesh including a neck. This is clearly highlighted by the systematic error around the neck region in Fig.~\ref{fig:head_eval}. In contrast, \modelnameHead captures the head deformations and the neck deformations plausibly and uniformly generalizes better.
\subsection{Hand Evaluation}
\label{hand_eval}
We use the publicly available \mano test set \cite{MANO:SIGGRAPHASIA:2017}. Since both SUPR-Hand and MANO share the same topology, we used the MANO test registrations provided by the authors to evaluate both models. To evaluate \ghumHand, we register the model to the MANO test set. However, the GHUM-Hand features a hand and an entire forearm, therefore to register GHUM-Hand we selected vertices on the model corresponding to the hand and only register that hand part of the model to the MANO scans. We fit all models to the corresponding registrations using a standard \emph{v2v} loss. For GHUM-Hand, we fit the model only to the selected hand vertices. The optimization free variables are the model pose and shape parameters. Fig.~\ref{fig:hand_num} shows generalization as a function of the number of shape parameters, where \modelnameHand uniformly exhibits a lower error compared to both \mano and \ghumHand. A sample qualitative evaluation of \mano and \modelnameHand is shown in Fig.~\ref{fig:hand_eval}. 
In addition to a lower overall fitting error, \modelnameHand has a lower error around the wrist region than \mano.
\begin{figure}[t]
         \begin{subfigure}[b]{0.80\textwidth}
         \vspace{-0.4cm}
         \includegraphics[width=\textwidth]{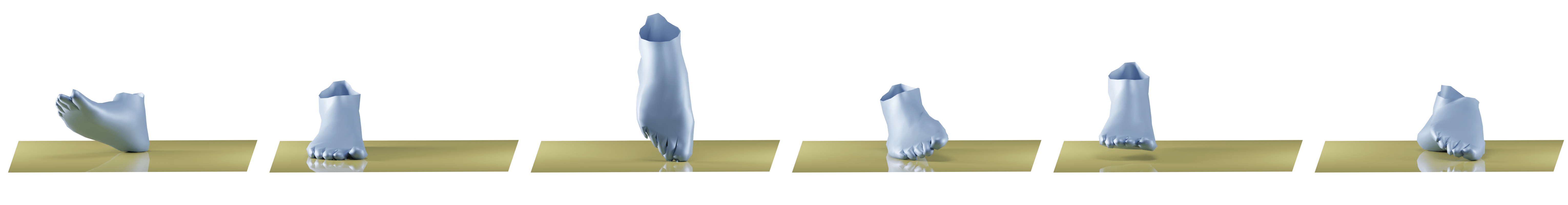}
            \caption{Registrations}
         \label{fig:body_rock}
     \end{subfigure}
    \begin{subfigure}[b]{\textwidth}
                  \begin{subfigure}[b]{0.80\textwidth}
             \includegraphics[width=\textwidth]{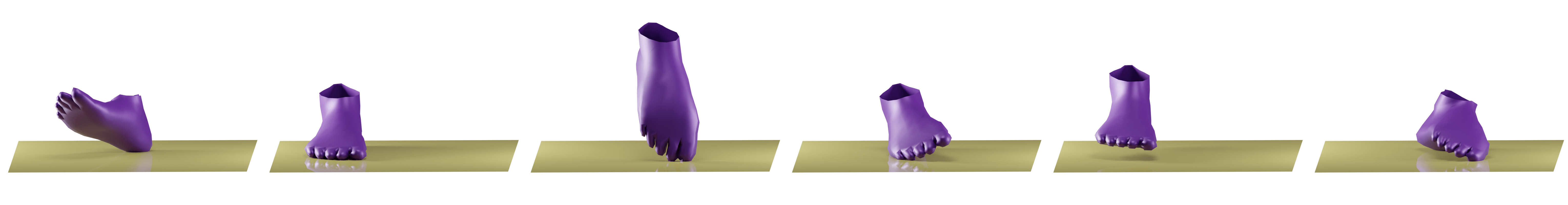}
             \includegraphics[width=\textwidth]{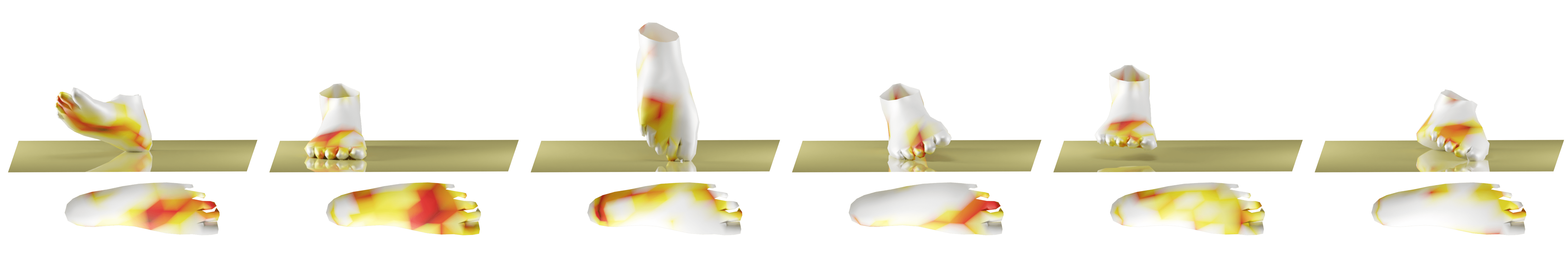}
             \caption{SMPL-X-Foot}
             \label{fig:body_rock}
    \end{subfigure}
    \begin{subfigure}[b]{0.02\textwidth}
              \includegraphics[width=1.0cm,height=1.5in]{CR_Figures/FIG_Bar.pdf}
     \end{subfigure}
   \end{subfigure}
       \begin{subfigure}[b]{0.80\textwidth}
         \includegraphics[width=\textwidth]{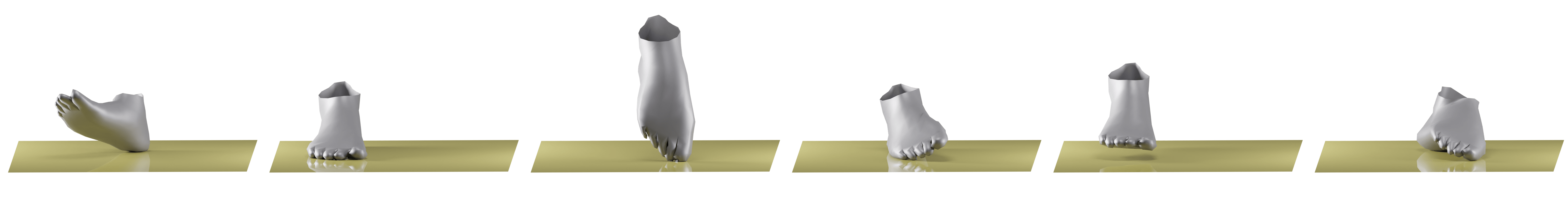}
         \includegraphics[width=\textwidth]{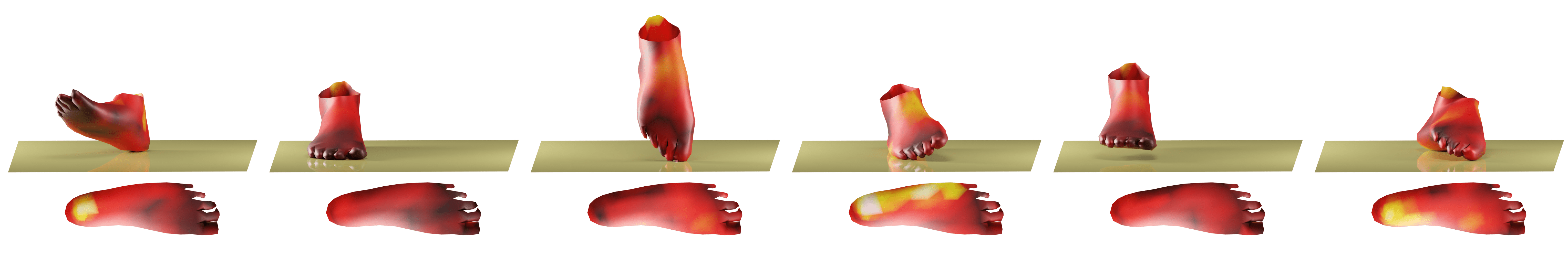}
         \caption{\modelnameFoot}
         \label{fig:body_rock}
     \end{subfigure}
     \vspace{-0.1in}
    \caption{Evaluating \modelnameFoot against SMPL-X-Foot.}
    \label{fig:foot_qualt_eval}
\end{figure}

\subsection{Foot Evaluation}
\label{foot_eval}

We evaluate \modelnameFoot generalization on a test set of held-out subjects. The test set contains $120$ registrations for $5$ subjects that explore the foot's full range  of motion, such as ankle and toe movements. 
We extract the foot from the SMPL-X body model as a baseline and refer to it as SMPL-X-Foot. 
We register the \modelnameFoot template to the test scans and fit the \modelnameFoot and SMPL-X-Foot to the registrations using a standard \emph{v2v} objective. For \modelnameFoot, the optimization free variables are the model pose and shape parameters, while for SMPL-X-Foot the optimization free variables are the foot joints and the SMPL-X shape parameters. We report the models' generalization as a function of the number of shape components in  Fig.~\ref{fig:foot_num}. A sample of the model fits are shown in Fig.~\ref{fig:foot_qualt_eval}. \modelnameFoot better captures the degrees of freedom of the foot, such as moving the ankle, curling the toes, and contact deformations.

\subsubsection{Dynamic Evaluation} 
\begin{figure}[t]
        \centering
         \begin{subfigure}[b]{0.80\textwidth}
         \centering
         \includegraphics[width=\textwidth]{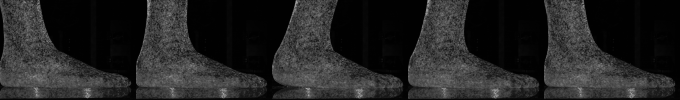}
         \caption{Raw Scanner Images}
         \label{fig:body_rock}
        \end{subfigure}
        \begin{subfigure}[b]{0.80\textwidth}
                 \includegraphics[width=\textwidth]{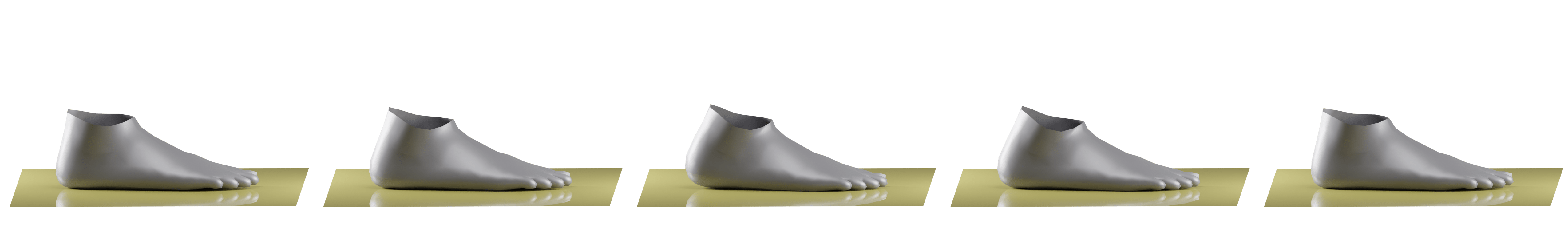}
                    \includegraphics[width=\textwidth]{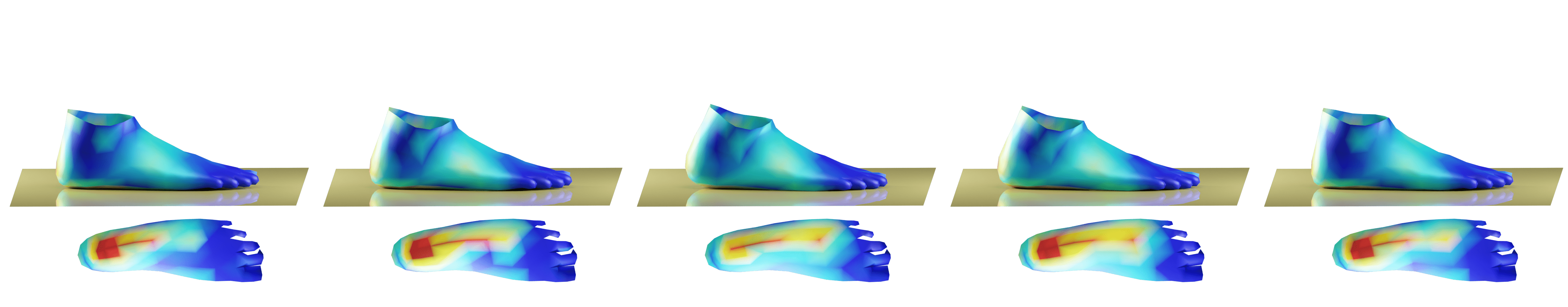}
                    \caption{\modelnameFoot predicted deformations}
                    \label{fig:foot_deform}
        \end{subfigure}
                \begin{subfigure}[b]{0.1\textwidth}
                    \includegraphics[width=0.8cm,height=1.5in]{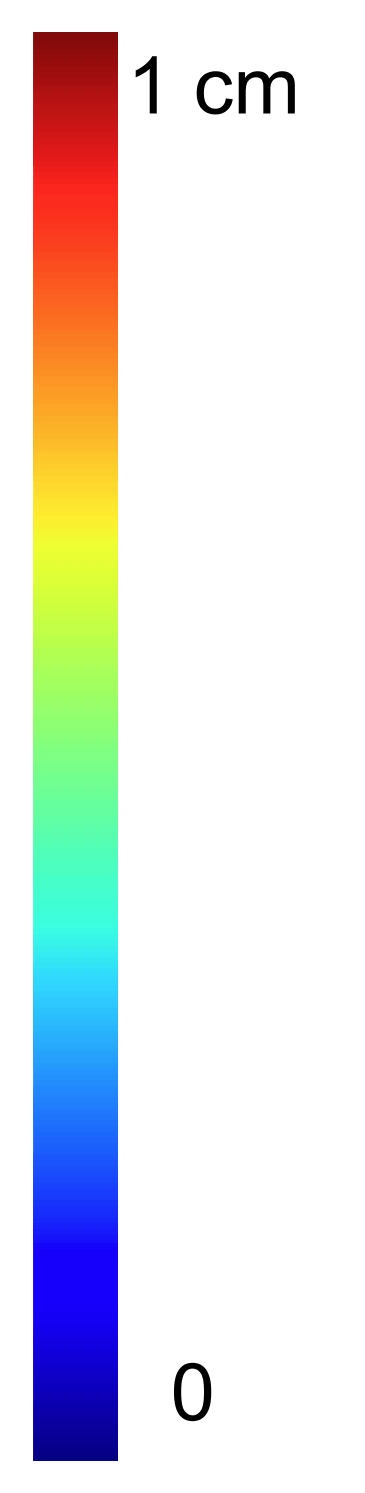}
                \end{subfigure}
             \begin{subfigure}[b]{0.80\textwidth}
         \centering
         \includegraphics[width=\textwidth]{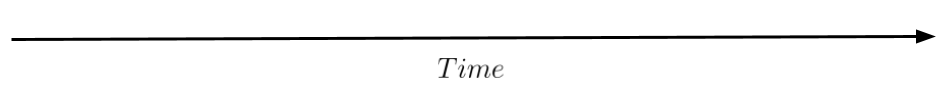}
     \end{subfigure}
    \caption{\textbf{Dynamic Evaluation}: Evaluating the \modelnameFoot predicted deformations on a dynamic sequence where the subject leans backward and forward, effectively shifting their center of mass. }
    \label{fig:dynamic_evaluation}
\end{figure}

We further evaluate the foot deformation network on a dynamic sequence shown in Fig.~\ref{fig:dynamic_evaluation}. Fig.~\ref{fig:body_rock} shows raw scanner footage of a subject performing a body rocking movement, where they lean forward then backward effectively changing the body center of mass. We visualise the corresponding \modelnameFoot fits and a heat map of the magnitude of predicted deformations in Fig.~\ref{fig:foot_deform}. When the subject is leaning backward and the center of mass is directly above the ankle, the soft tissue at heel region of the foot deforms due to contact. The \modelnameFoot network predicts significant deformations localised around the heel region compared to the rest of the foot. However, when the subject leans forward the center of mass is above the toes, consequently the soft tissue at the heel is less compressed. The \modelnameFoot predicted deformations shift from the heel towards the front of the foot. 

\section{Conclusion}
\label{sec:conclusion}
We present a novel training algorithm for jointly learning high-fidelity expressive full-body and body parts models. We highlight a critical drawback in existing body part models such as FLAME and MANO, which fail to model the full range of motion of the head/hand. 
We identify that the issue stems from the current practice in which body parts are modeled with a simplified kinematic tree in isolation from the body. 
Alternatively, we propose a holistic approach where the body and body parts are jointly trained on a federated dataset that contains the body parts' full range of motion relative to the body. 
Additionally, we point out the lack of any articulated foot model in the literature and 
show that the feet of existing full-body models do not have enough joints to model the full range of motion of the foot. 
Using 4D scans, we learn a foot model with a novel pose-corrective deformation formulation that is conditioned on the foot pose, its shape, and ground contact information. 
We train \modelname with a federated dataset of $1.2$ million scans of the body, hands, and feet. 
The sparse formulation of \modelname enables separating the model into an entire suite of body-part models. 
Surprisingly, we show that the head and hand models are influenced by significantly more joints than commonly used in existing models. 
We thoroughly compare \modelname and the separated models against SMPL-X, \ghum, MANO and FLAME and show that the  models uniformly generalize better and have a significantly lower error when fitting test data.  
The pose-corrective blendshapes of \modelname and the separated body part models are linearly related to the kinematic tree pose parameters, therefore our new formulation is fully compatible with the existing animation and gaming industry standards. A Tensorflow and PyTorch implementation of \modelname and the separated head  (\modelnameHead),  hand (\modelnameHand) and the foot (\modelnameFoot) models is publicly available for research purposes. 

\textbf{Acknowledgments:} The authors thank the MPI-IS Tübingen members of the data capture team since $2012$ for capturing the data used to train SUPR: S. Polikovsky,  
A. Keller,
E. Holderness,
J. Márquez,
T. Alexiadis,
M. Höschle,
M. Landry,
G. Henz,
M. Safroshkin,
M. Landry,
T. McConnell, T. Bauch and B. Pellkofer for the IT support. The authors thank M. Safroshkin and M. Landry for configuring the foot scanner. The authors thank the International Max Planck Research School for Intelligent Systems (IMPRS-IS) for supporting Ahmed A. A. Osman. This work was done when DT was at MPI. 
\par 
\textbf{MJB Disclosure:} \url{https://files.is.tue.mpg.de/black/CoI_ECCV_2022.txt}

\bibliography{egbib}
\clearpage 


\section*{Supplementary material}

# SUPR: A Sparse Unified Part-Based Human Representation
## *** Supplementary Material ***

Ahmed A. A. Osman[1], Timo Bolkart[1], Dimitrios Tzionas[2], and
Michael J. Black[1]

[1] Max Planck Institute for Intelligent Systems, Tübingen, Germany
[2] University of Amsterdam
{aosman,tbolkart,black}@tuebingen.mpg.de,d.tzionas@uva.nl

## 1  Overview

*Paper Summary.* We highlight drawbacks in widely used body part models. Existing head/hand models do not model the head/hand full range of motion. We address this with a new holistic learning scheme in which we train body parts models jointly with the body. To this end, we train SUPR an expressive human body model, where the each joint strictly influences a sparse set of the model vertices. This sparse factorization enables us to separate SUPR into a full suite of high fidelity body part models. We show that body part hand/head models learned jointly with the body are influenced by significantly more joints than existing part models. Additionally, we note that, despite the importance of the foot for locomotion, there is no existing foot part model and the feet of full body models are significantly under actuated. To address this, we learn a foot model from novel 4D scans and propose a novel function that relates the foot deformation to the foot shape, pose and ground contact.

### 1.1  Paper Content

In this Supplementary Material, we perform extensive ablation studies and evaluations to further explore the main paper's key contributions. The rest of the document is arranged as follows: in Section 2 we describe the federated training dataset of scans, including the foot scanner that enables us to model the foot deformations due to contact. The foot deformation architecture is described in detail in Section 3. We describe the SUPR training in Section 4. In Section 5 we further evaluate SUPR. Ablations for the SUPR-Foot network are introduced in Section 6. As introduced in the main paper, SUPR is based on spherical joints, which produce redundant degrees of freedom (DoF) for joints like those of the fingers. In Section 7 we describe a constrained version of the kinematic tree with fewer DoF. We provide a comparison between SUPR and existing expressive human body models in Section 8. We conclude by discussing limitations of our method in Section 9

## 2    Data

SUPR is trained on a federated dataset of 3D scans. In total 4 types of scanners are used: a full body scanner, a hand scanner, a head scanner and a foot scanner. All the scanners are 4D scanners, capturing high resolution dynamic sequences for each body part. We additionally leverage datasets that are either publicly available for research purposes or commercial datasets from private vendors. In this section we describe the scanning setup for each scanner and describe the external datasets. We discuss the foot scanner in Section 2.5 which was key to enable us to capture the foot including the toes and the foot soles. In the main paper we highlight that SUPR and the separated body part models were trained on 1.2 millions scans, a break down of the number of scans for each body part is discussed in Section 2.6.

### 2.1    Body Scanner

Human bodies deform in complex ways as a result of changes in body pose and body shape. To study and model minimally-clothed human body deformations, we use a 4D scanner that captures the full 3D human body shape at 60 frames per second (fps). The full-body scanner is custom built by 3dMD (Atlanta, GA). The system uses 22 pairs of stereo cameras, 22 color cameras, and speckle-light projectors. The speckle patterns allow accurate stereo reconstruction of 3D shape. This speckle pattern alternates at 120 fps with large white-light LED panels that provide a smooth nearly uniform illumination. The scanner outputs high resolution meshes with approximately 150,000 vertices. The high resolution meshes in addition to the high frame rate (60 fps) enable us to model the subtle deformations of the human body. The full body scanner scanning volume is sufficient to capture poses such as a full leg split by a ballerina, or a sitting or lying down poses.

The captured data contains a wide diversity of body shapes. The training scans include extreme body shapes such as body builders and anorexia nervosa patients. Furthermore, the data capture protocol include athletes such as a ballerina and a yoga expert. Additionally, since SUPR has a full expressive kinematic tree, including a fully articulated hand, jaw and an expressive head, we capture expressive sequences where subjects performed motions communicating emotions and intent. An overview of the full body training scans is shown in Figure 1

**External datasets**    In addition to the scans from the 4D body scanner, we leverage a number of datasets of 3D human body scans. To capture the diversity of human body shape we use the CAESAR [10] and SizeUSA [1] datasets. The CAESAR database contains 1700 male and 2107 female subjects distributed according to the US population in 1990. A limitation of CAESAR's capture protocol is that all women subject were in sports-bra-type top. As a result of the bra type, the CAESAR female chest shape does not reflect the diversity

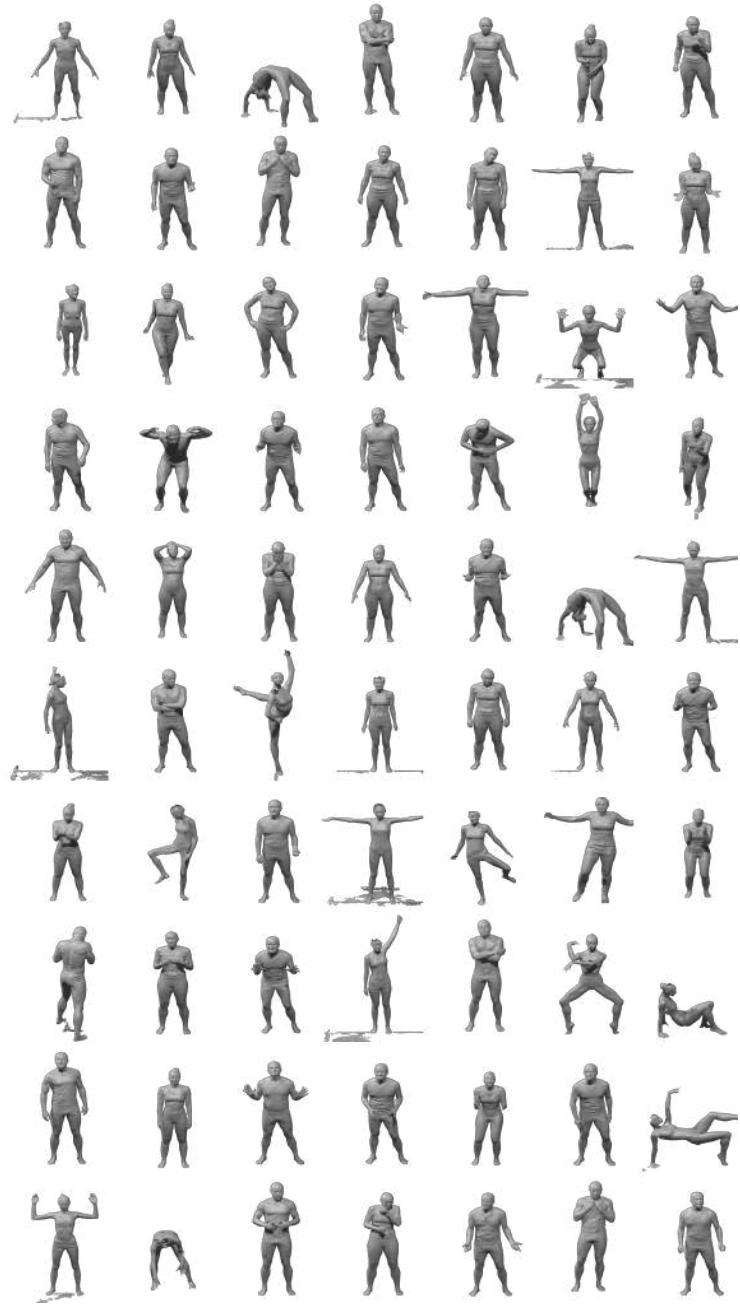

Fig. 1: Overview of the scans captured in the full body scanner. The scans are detailed and high resolution. Note however, the hands and the feet are poorly reconstructed, and the head resolution is not sufficient to capture subtle facial expressions.

of shapes found in real applications. We additionally use the SizeUSA dataset, which contains a richer diversity of body shapes and the female subjects wore a traditional bra. The SizeUSA dataset contains $10,000$ subjects ($2845$ male and $6436$ female).

Despite the 4D scanner high resolution output meshes, the output scans have have poorly reconstructed hands and foot. The foot sole is poorly reconstructed because it is always occluded by the glass platform. The full body scans are not suitable for learning head, hand and foot deformations.

## 2.2   Head Scans

The human head exhibits a range of highly dynamic deformations. When we refer to the head we mean the face, the back of the head including the scalp and the neck. The human head 3D deformations are due to facial expressions, jaw movement, head movement relative to the neck and body movement relative to the neck (for example when shrugging). We use a dedicated head scanner to complement the full body 4D scanner. The head scanner has a significantly higher number of cameras focused on the head region compared to the body scanner in Section 2.1. The scanning setup enables us to capture the subtle facial expressions. We note, however, that the head scanner has a limited scanning volume making it infeasible to capture the full range of motion of the human head relative to the body.

Similar to the full body scanner, the head scanner is a 4D scanner capturing high-resolution dynamic sequences. The scanner employs 6 pairs of stereo cameras to compute shape and geometry with the assistance of custom speckle projectors. It also includes 6 color cameras and white-light panels to capture texture. The data capturing protocol was designed by experts to capture subtle and extreme facial expressions, full movement of the jaw, in addition to neck movement poses such as looking up, down to the left or right.

## 2.3   Hand Scans

The reconstructed fingers in full-body scans are typically noisy and poorly reconstructed, as shown in Figure 1. To better capture the hands, we use the data from the MANO hand model [11]. These hand scans are used to learn the pose corrective blend shapes due to finger articulation. A sample of the captured hand scans is shown in Figure 3.

## 2.4   Foot Scanner

The human foot is a complex structure containing muscles, other soft tissue, and a quarter of the bones in the human skeleton. All existing human body models  [2,8,7,9,12,5] use a highly simplified kinematic tree to model the feet with a limited number of joints. Such modes can not model the full range of

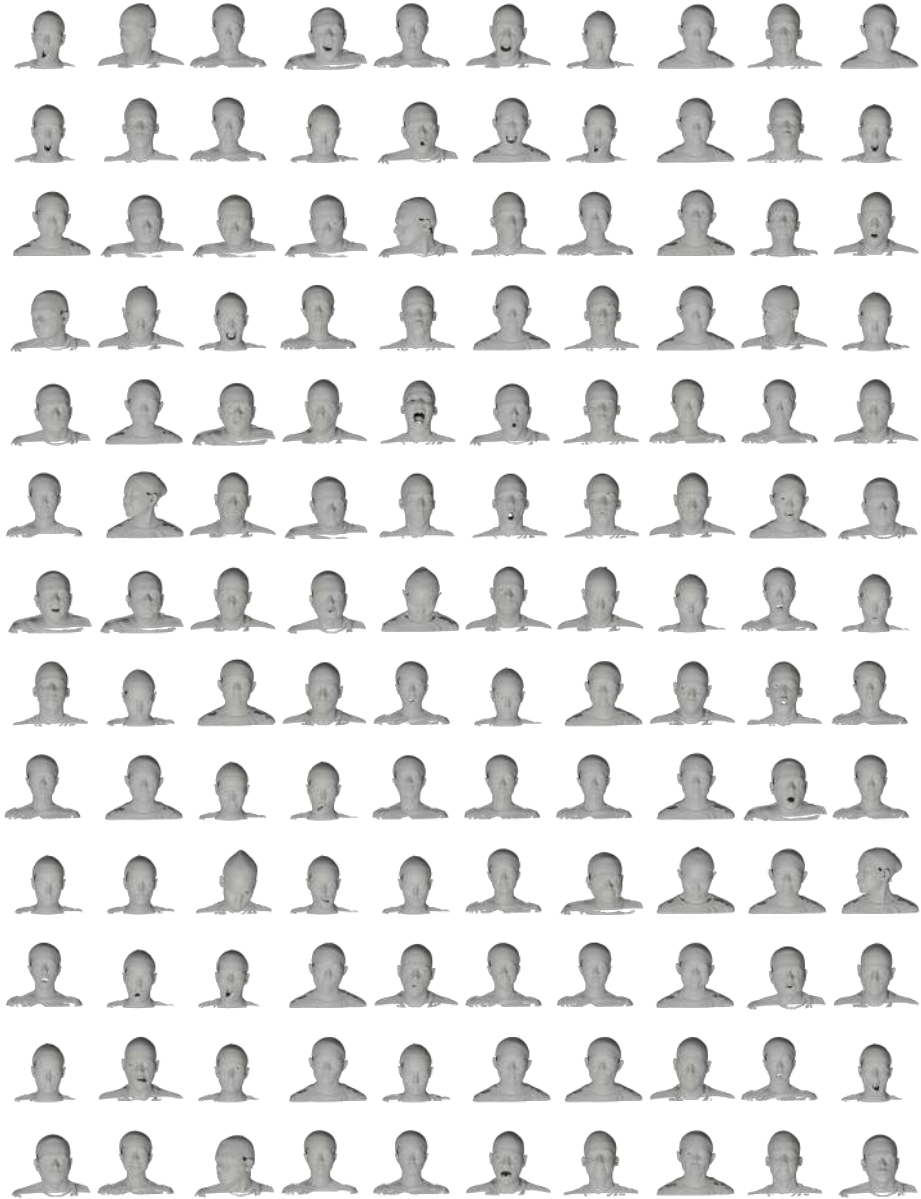

Fig. 2: **Head Scans** A sample of the head scans using in training SUPR.

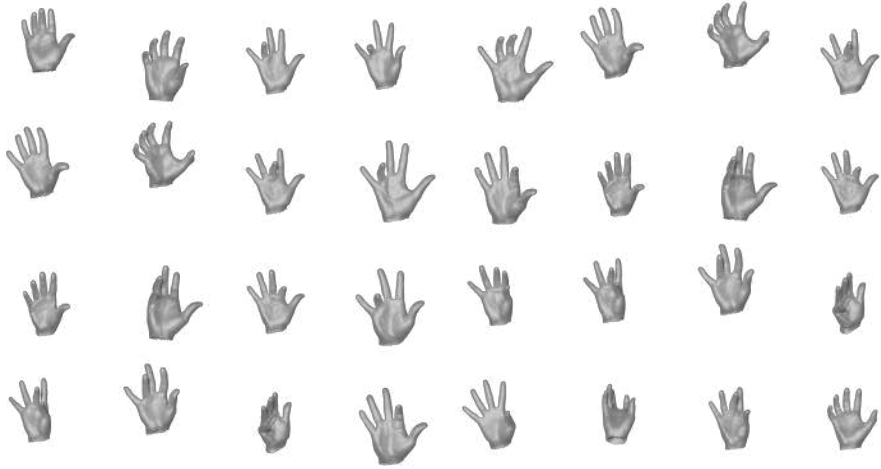

Fig. 3: **Hand Scans:** A sample of the hand scans used to train SUPR.

motion of the bones in the human foot, as highlighted in Figure 3 in the main paper. The kinematic tree of models like SMPL-X is not sufficient to capture the toe articulation. The commonly used pose deformation functions only define body deformations due to pose. As we discuss in the introduction section in the main paper, this is insufficient to capture the foot deformations due to ground contact.

**Existing Foot Models.** The key reason existing models fail to accurately model foot deformations is because the foot dynamic deformations are hard to capture. The only interesting exception is the work of Boppana et al. [3] discussed in Section 2 in the main paper, which is, to best of our knowledge, the first and only work that attempts to build a model of dynamic foot deformations. Bopanna et al. recognize the limitation of existing scanning solutions to model the human foot deformations and propose the DynaMo system. The Dynamo scanning setup is comprised of a treadmill surrounded by 7 Intel RealSense cameras. A total of 30 subjects were captured walking on a treadmill. A sample reconstructed mesh by the DynaMo system is shown in Figure 6b. We note here that the output scan is noisy and does not capture the foot sole. Bopanna et al. register the DynaMo scans to a high resolution template mesh and learns a PCA space on the registered meshes. The model they propose does not contain toes or a foot sole as shown in Figure 6. This is not surprising given the fidelity of the foot scans captured by the DynaMo system. In contrast to the Bopanna et al. model, SUPR-Foot contains an extensive kinematic tree with 13 joints per foot as shown in Figure 3c in the main paper. SUPR is the first articulated model of the human foot with a pose space that can be driven by bio-mechanics simulations of the

human foot for example, in addition to a shape space to capture the diversity of human foot shape.

## 2.5   Foot Scanner

SUPR goes beyond existing expressive human body models to model the human foot. To enable capturing the full range of the human foot deformations, we use a custom built scanner dedicated for the foot. The scanner is designed to be mechanically stable to capture dynamic poses such as walking, running or jumping. The output scans are high resolution and can capture the movement of the toes. The scanner floor is a transparent glass platform (which can support subjects up to 150 kg), which enables us to capture the foot sole deformation due to ground contact.

An overview of the foot scanner is shown in Figure 5. The scanner setup features a runway for the subjects to run or walk. In Figure 5b, we show raw scanner images, where the foot is visible from all views, including the foot sole. The scanner uses 10 pairs of stereo cameras, including dedicated cameras capturing the bottom of the foot. The frame rate of the scanner is 10 fps. The output scans contain on average 30,000 points.

**Data Capture Protocol.** We capture a total of 30 subjects, 15 female and 15 male subjects with a total of 70,000 scans. The data capture protocol is designed by experts to explore the space of human foot deformations. The capture protocol is divided into two main parts: 1) Non-Contact sequences 2) Contact Sequences. In the non-contact sequences, the subject foot is not in contact with the glass platform. The data capture protocol for such sequences is designed to explore the full degree of freedom of the toes and the ankle. In contact sequences the subject's foot is partially or in full contact with the glass platform. The contact sequences include motions such as walking/running and jumping. In total We capture 356 dynamic sequences which is the largest training dataset for human scans report in the literature. An overview of the captured scans is shown in Figure 4

**Foot Shape Scans.** The 30 subjects captured in the dynamic foot scanner do not represent the diversity of human foot shape. Accurate modeling of the human foot shape is crucial for the footwear industry. The feet in the CAESAR and SizeUSA scans, shown in Figure 7a, are noisy, missing, and are not good enough to learn a statistical model. To accurately model the diversity of the human foot scans, we acquired an additional 7,000 high resolution foot scans from the ANSUR II dataset [4]. Figure 7 compares the curated high resolution foot scans in comparison to CAESAR and SizeUSA foot scans. In contrast to CAESAR and SizeUSA, the curated dataset of foot scans is significantly less noisy, with on average 10x the resolution of a foot scans from CAESAR/SizeUSA. The high resolution foot scans preserve the 3D geometry of the individual toes. We use this data in learning the the local shape space of SUPR-Foot.

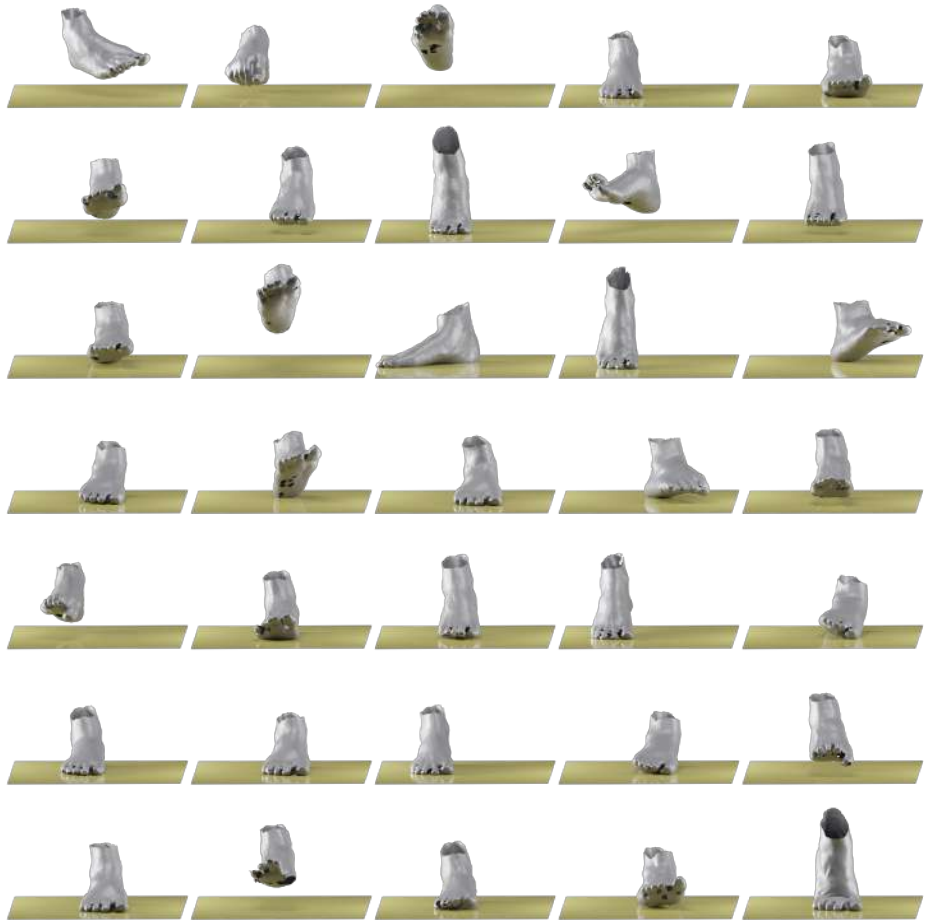

Fig. 4: **Foot Scans:** An Overview of the foot scans. The foot is full reconstructed including the toes and the foot sole.

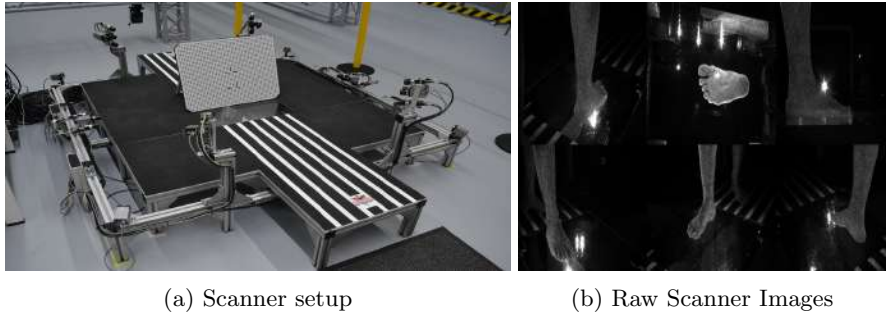

(a) Scanner setup                    (b) Raw Scanner Images

Fig. 5: A 3dmD foot scanner using 10 pairs of stereo, including dedicated cameras capturing the bottom of the foot through a transparent glass platform. The scanner features a run way to capture dynamic sequences such as walking.

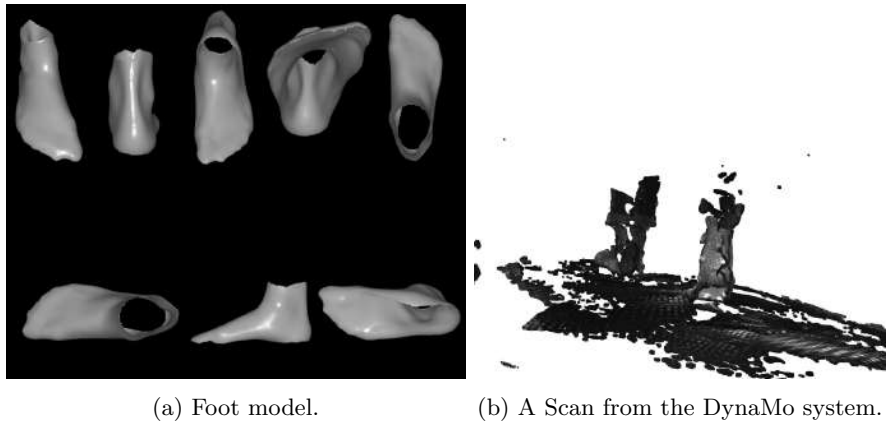

(a) Foot model.                    (b) A Scan from the DynaMo system.

Fig. 6: Bopanna et al. [3] foot model based on Principal Component Analysis of dynamic foot scans. The model does not contains toes or a foot sole.

### 2.6   Training Scans Summary

SUPR and the separated body parts are trained on a total of 1.2 million scans. In Figure 8 we compare the scale of training datasets used to train body models in the literature. As Figure 8 highlights, the scale of the training data is an order of magnitude larger than the largest training dataset reported in the literature (60K, for the GHUM model). A breakdown of the number of scans captured by each body part is summarized in Table 1.

## 3   SUPR-Foot Network

In the main paper Section 3.3 we introduce the foot deformation function. A key contribution of our paper is the deformation function relating the foot deforma-

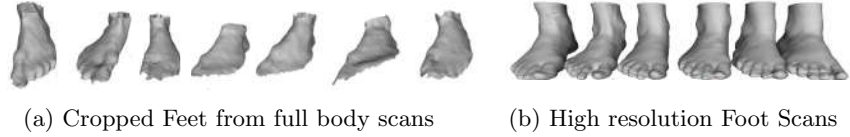

(a) Cropped Feet from full body scans          (b) High resolution Foot Scans

Fig. 7: Comparing reconstructed feet from a full body scanner compared to the curated high resolution foot scans. We curate a total of $7,000$ high resolution foot scans from the ANSUR II dataset [4]. The curate scans have 10x the resolution of foot scans captured in a body scanner and preserve the individual toes geometry.

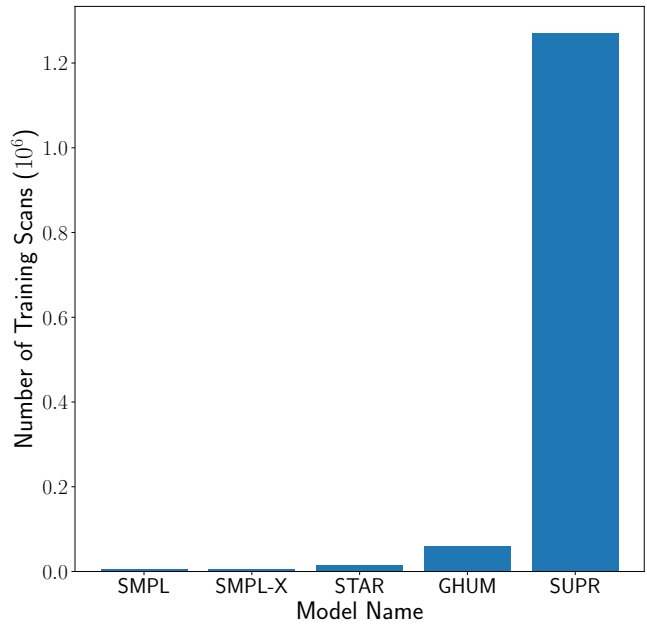

Fig. 8: A comparison between the scale of training datasets for recent human body models. SUPR is trained on a order of magnitude more data compared to the highest number of training scans report in the literature (GHUM 60k).

| Body Part | Number of Scans |
|-----------|-----------------|
| Body Scans | 775,481 |
| Head Scans | 421,898 |
| Foot Scans | 69,257 |
| Hand Scans | 3,531 |
| Total | 1,270,167 |

Table 1: A breakdown of the number of scans for each body part.

tion to the foot pose, shape and contact parameters. In this section we describe the foot deformation network in detail.

**Foot Contact** The raw foot scans generated by the the foot scanner described in Section 2.5 does not provide per vertex contact labeling, describing whether a vertex in the scan is in contact with the glass platform. To estimate the per-vertex ground contact information we register all the scans to the foot template mesh $\overline{T}_{foot}$. We additionally estimate the ground plane for each dynamic sequence by fitting a plane to the glass platform scan points. A vertex $\vec{v} \in \overline{T}_{foot}$ is labelled in contact with the ground, if it the point-to-plane distance between the vertex and the ground plan is less than a threshold. We allow for a soft threshold when estimating contact since the scans have noise. The threshold used in the main paper is 0.1 mm.

## 3.1   Foot Deformation Network

The foot deformation network is an encoder-decoder architecture as described in Section 3.3 in the main paper. We train a deformation network for each foot separately. Below we describe the network for the right foot. We first introduce the notation we use:

- $B_P$: is the linear pose corrective blend shape described Equation 1 in the main paper.
- $B_C$: is the predicted deformations for the foot related to pose, contact and foot shape.
- $\vec{c}$: is a binary vector of which vertices are in contact with the glass platform.
- $\vec{z}$: is a latent code vector.
- $\vec{\theta}$: is a foot pose parameters.
- $\vec{\beta}$: is a foot shape parameters.
- $\vec{f}$: is a concatenated feature of the pose, shape and contact vector.
- LReLU: leaky rectified linear units with a slope of 0.1 for negative values.
- $FC_m$: fully connected layer with output dimension $m$.

**Feature Representation** The input to the network $\vec{f}$ is a concatenation feature representation of the foot pose, foot shape and contact. The foot pose representation is based on normalized unit quaternion representation defined by:

$$F(\vec{\theta}) = Q(\vec{\theta}) - Q(\vec{\theta^*}) \tag{1}$$

where $Q(.) : \mathbb{R}^3 \to \mathbb{R}^4$ is a function computing the quaterion representation of the input axis angle rotation, $\theta^*$ is the foot in the rest pose. The feature representation in Equation 1 will evaluate to 0 when the foot is in the rest pose. The foot template mesh $\overline{T}_{foot} \in \mathbb{R}^{266 \times 3}$ is a high dimensional representation to represent the foot shape. We represent the foot shape using the first two principal components which correspond to the foot length and foot volume. We experimented with different number of coefficients, and the first two PC component result in the lowest generalization error on the validation set. The state of the foot contact with the scene is represented using the $\vec{c}$. More formally the input feature to our network:

$$\{F(\vec{\theta}), \vec{\beta}_1, \vec{\beta}_2, \vec{c}\} \xrightarrow{\text{concat}} \vec{f} \in \mathbb{R}^{320}, \tag{2}$$

where $\vec{\beta}_1, \vec{\beta}_2$ are the first two PCA components and the *concat* operator is a standard vector concatenation operator.

### 3.2 Architecture

The architecture is an encoder-decoder fully-connected network, with non-linear activations based on LReLUs. Encoder:

$$\vec{f} \in \mathbb{R}^{320} \to FC_{256} \to$$
$$\to FC_{128} \to FC_{64} \to$$
$$\to FC_{32} \to \vec{z} \in \mathbb{R}^{16}$$

The dimensionality of the latent code $\vec{z}$ was chosen by grid search. We experimented with dimensionality 64, 32 and 16. A latent code with dimensionality 16 result in the lowest generalization error of the validation set. The decoder is described by:

$$\vec{z} \in \mathbb{R}^{16} \to FC_{32} \to$$
$$\to FC_{64} \to FC_{128} \to FC_{266} \to B_C$$

where $B_C$ is added to the linear blend shape $B_P$ as shown in Equation 6 in the main paper.

# 4 Training

## 4.1 SUPR Training

SUPR pose corrective formulation is training is similar to STAR. The key difference is the pose corrective formulation of SUPR is not conditioned on body shape similar to STAR. The additional shape dependant blend shape is not sparse. The key reason we are able to separate SUPR is the fully sparse factorization of the pose blend shapes and the skinning weights as discussed in Section 3 in the main paper.

The SUPR pose corrective blend shapes are trained by minimizing the reconstruction loss between between the model prediction and the federated dataset of groundtruth registration. The SUPR pose blend shape parameter, namely the joint activation $\mathcal{A}$ and the pose corrective blend shapes $\mathcal{P}$ are trained by stochastic gradient descent. Since our data is based on 4D dynamic sequences, we first shuffle the data such that there is no similarity between subsequent frames. We use batches of size 32 to minimize the vertex-to-vertex loss given by:

$$\mathcal{L}_D = \frac{1}{B} \sum_{i=1}^{32} ||M(\theta^i) - R^i||_2. \tag{3}$$

where $R^i$ is the ith groundtruth registration in the batch. Similar to STAR, we use an $L1$ penalty on the output of the joint activation $\mathcal{A}$,

$$\mathcal{L}_A = \lambda_c || \sum_{i=1}^{K-1} \phi_j(A_j) ||, \tag{4}$$

where $\lambda_c$ is a scalar constant. The full objective for the pose space is defined by Equation:

$$\mathcal{L} = \mathcal{L}_A + \mathcal{L}_D, \tag{5}$$

where Equation 5 is minimized with respect to the pose corrective regression weights $\mathbf{K_{1:80}}$, activation weights $\mathbf{A_{1:80}}$. We use a batch size $B = 32$ and the ADAM.

## 4.2 Body Part Training

Given the trained blend shapes, the body part pose space is separated as discussed in Section 3 in the main paper. We further train a local shape space for each of the separated body part models. The CAESAR head, hand of the CAESAR registrations are used to train a local shape space for SUPR-Head and SUPR-Head. The local shape space for the foot is trained on the curate high resolution foot scans, as the foot in the CAESAR scans were noisy. The percentage of explained variance as the number of shape components for each body part is shown in Figure 9.

### 4.3   Deformation Network Training

Given the learned linear blend shapes trained in Section 4.1, and the local shape space for the foot, we train the deformation network for the foot deformation described in Section 3 in the main paper. For training the deformation network we use both contact and non-contact foot registrations. The network is trained by minimizing the $L1$ loss between the model and the foot registrations:

$$\mathcal{L} = ||M(\theta^i, \vec{c}^i, \vec{\beta}^i) - R^i||. \tag{6}$$

The training loss is minized using stochastic gradient descent, where we used ADAM with batch size 32.

## 5   SUPR Ablation

### 5.1   STAR Evaluation

SUPR pose corrective blend shape formulation is based on the STAR pose corrective blend shape formulation as discussed in Section 3 in the main paper. For completeness we further evaluate SUPR against STAR. We note however that SUPR is expressive, with 1.5x more vertices and 3x more joints compared to STAR. We use the 3DBodyTex dataset and register the scans to the STAR template. A human expert validated all registrations. Similar to the evaluation Section 4.1, we fit each model by minimizing the vertex-to-vertex loss ($v2v$) between the model surface and the corresponding registration. The free optimization parameters for both models are the pose parameters $\vec{\theta}$ and the shape parameters $\vec{\beta}$. We report the model generalization error in Figure 10.

### 5.2   GHUM Body Parts Evaluation

Head Evaluation In Section 4.2 in the main paper we evaluate SUPR-Head against GHUM-Head. A sample qualitative comparison fits are shown in Fig. 12. Similar to FLAME, GHUM-Head has significant error around the neck region.

**Hand Evaluation** In Section 4.3 in the main paper we evaluate SUPR-Hand, against GHUM-Hand. A sample of the GHUM-Hand fits are shown in Figure 11. GHUM-Hand consistently has a high error in the wrist region compared to the the fingers.

## 6   SUPR-Foot Ablation

In the main paper Section 4.4 we evaluate SUPR-Foot against SMPL-X-Foot on a held out test set of contact and non-contact foot scans. We further break down the evaluation in Figure 13. We report the model mean absolute error as a function of the number of shape components used on non-contact frames in Figure 13a and contact frames in Figure 13b.

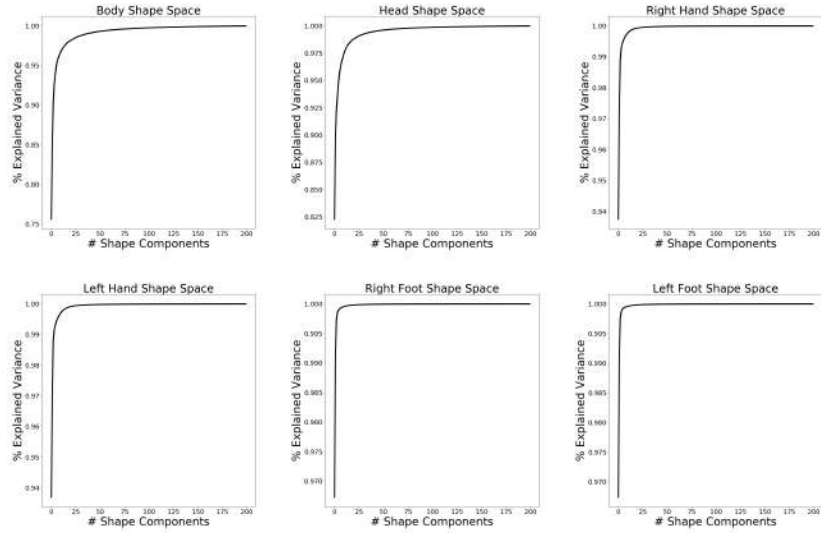

(a) Male Shape Space

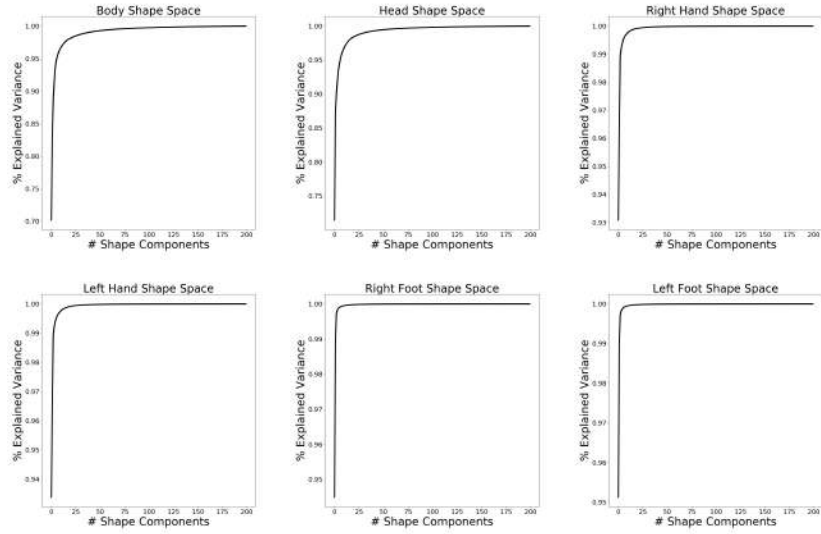

(b) Female Shape Space

Fig. 9: Percentage of explained variance as a function of the number of shape components for SUPR and the separated body part models.

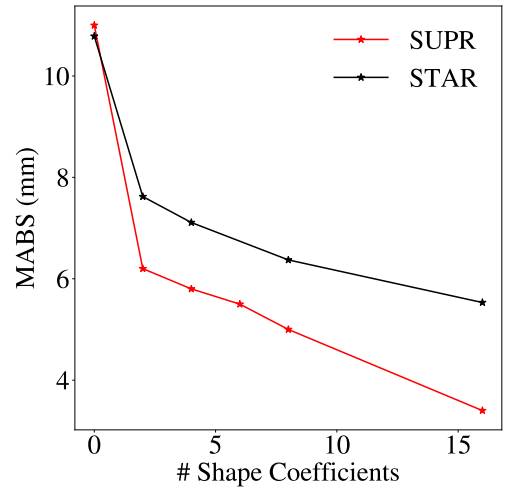

Fig. 10: Evaluation of SUPR against STAR on the 3DBodyTex dataset.

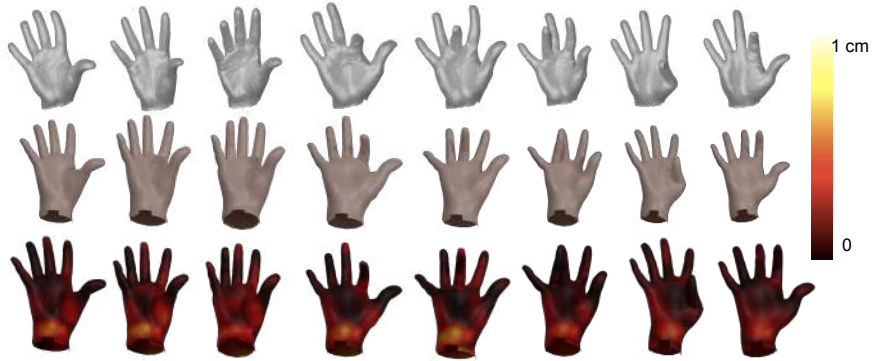

Fig. 11: **GHUM-Hand Evaluation :** Evaluating GHUM-Hand on the MANO test set. The top row are raw scans from the MANO test set, the second row is GHUM-Hand model fits with 10 shape components, while the bottom row is the corresponding error heatmap.

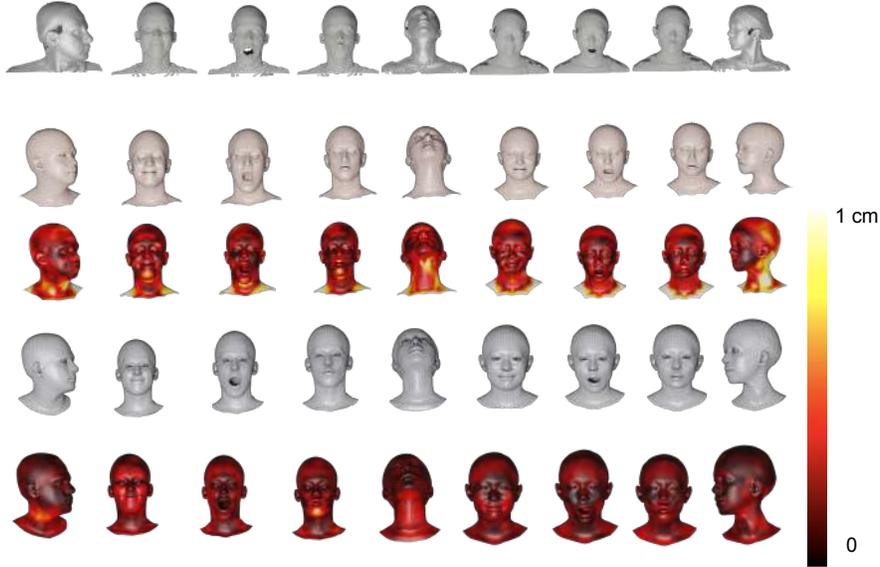

Fig. 12: **Head Evaluation:** Evaluating GHUM-Head and SUPR on a test containing head movement relative to the neck, jaw movement and extreme facial expressions. Both GHUM-Head and SUPR are fit with 16 shape and epxression components. The top raw correspond to head raw scans, second row correspond to GHUM-Head fits, third row correspond to the heatmap, fourth row correspond to SUPR-Head fits and the fifth row are the corresponding error heat maps.

Table 2: Foot Deformation Ablation Study. SUPR-Foot $_{lbs}$ corresponding to model with linear blend skinning, no additive correctives used. SUPR-Foot $_{lbs+l}$ correspond to lbs in addition to the linear correctives, SUPR-Foot $_{lbs+l+f(\theta)}$ is adding the non-linear deformation where the network is condition on pose only, SUPR-Foot $_{lbs+l+f(\theta,\vec{\beta})}$ where the network is conditioned on pose and shape information, while SUPR-Foot is the full model.

| Model | Non-Contact v2v (mm) ↓ | With-Contact v2v (mm) ↓ |
|---|---|---|
| SUPR-Foot $_{lbs}$ | 5.235 ±0.126 | 6.691 ±1.369 |
| SUPR-Foot $_{lbs+l}$ | 4.587 ±0.589 | 5.364 ±1.279 |
| SUPR-Foot $_{lbs+l+f(\theta)}$ | 2.982 ±0.859 | 4.129 ±1.883 |
| SUPR-Foot $_{lbs+l+f(\theta,\beta)}$ | 2.910 ±0.728 | 3.934 ±1.819 |
| SUPR-Foot (ours) | 2.753 ±0.821 | 3.122 ±1.462 |

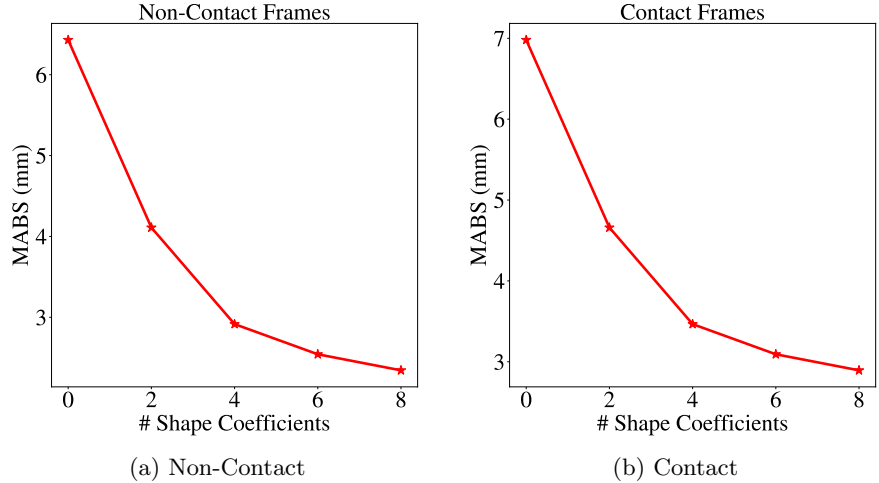

(a) Non-Contact              (b) Contact

Fig. 13: Evaluating SUPR-Foot on frames where the foot was not in contact with the glass platform shown in Figure 13a, and frames where the foot was partially or fully in contact with the glass platform in Figure 13b.

*Deformation function:* A key contribution of our work is introducing a novel deformation function which relates the foot deformations to the foot pose, shape and ground contact. We illustrate the influence of each term on the model generalization by ablating the foot deformation network described in Section 3 in the main paper. We retain variations of the deformation network from scratch and refit each model to the test set. We report the model $v2v$ error in Table. 2. The result clearly show the vertex to vertex error decreasing on the held out test set when adding each term in the foot deformation function across both the contact and non-contact frames.

## 7   Constrained SUPR

The SUPR kinematic tree introduced in Section 3 is based on spherical joints. Each spherical joint $j$ is parameterized by $\vec{\theta_j} \in \mathbb{R}^3$. The spherical joints allow redundant degrees of freedom for some body parts such as the fingers. For the fingers, for example, the axes of rotation are not bone-aligned. In order to simply bend a finger we have to control 3 axis-angle rotations. This is problematic to use by animators and for architectures that regress hand pose parameters from images. In this section we describe a constrained version of SUPR that uses hinge/double hinge joints in contrast to spherical joints.

**Constrained Kinematic Tree.** The kinematic tree of the constrained version of SUPR (shown in Figure 14) uses hinge and double hinge joints. A hinge joint

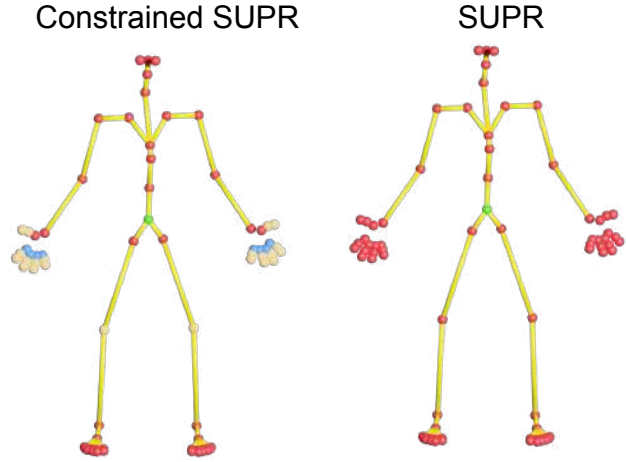

**Constrained SUPR**        **SUPR**

Fig. 14: **Constrained SUPR Kinematic Tree:** SUPR is based on spherical joints which allow redundant degrees of freedom for body parts such as the fingers. The constrained SUPR kinematic tree contains a mixture of joints: Spherical joints (shown in red), Hinge Joints (shown in beige) and double hinge joints (shown in blue).

is fully parameterized by an axis of rotation $\vec{a} \in \mathbb{R}^3$ and a pose parameter $\vec{\theta} \in \mathbb{R}$. A double hinge joint is defined by two axes of rotation and pose parameters $\vec{\theta} \in \mathbb{R}^2$. The axes of rotation for the hinge and double hinge joints are orthogonal to the bone. Therefore, to simply bend a finger in SUPR requires only controlling or regressing one or two scalars. This compact representation is convenient for artists, regression tasks and is more anatomically plausible.

Specifically, this version of SUPR is defined by Eq. 7:

$$M(\vec{\theta}, \vec{\beta}, \vec{\psi}) = W(T_p(\vec{\theta}, \vec{\beta}, \vec{\psi}), J(\vec{\beta}), AX, \vec{\theta}, \mathcal{W}), \tag{7}$$

where $AX \in \mathbb{R}^{30 \times 3}$ is the axis of rotation matrix for the hinge and double hinge joints. The key difference between Equation 1 in the main paper and Equation 7 is the bone transformation rotation matrix. The rotation matrix for a hinge joint is a constrained rotation matrix, which only allows a single degree of freedom with respect to the axis of rotation $\vec{a}$. A constrained rotation matrix is defined by:

$$\begin{bmatrix} a_x^2 + c_\theta(1-a_x^2) & a_x a_y(1-c_\theta) + a_z s_\theta & a_x a_z(1-c_\theta) - a_y s_\theta \\ a_x a_y(1-c_\theta) - a_z s_\theta & a_y^2 + c_\theta(1-a_y^2) & a_y a_z(1-c_\theta) + a_z s_\theta \\ a_x a_z(1-c_\theta) + a_y s_\theta & a_y a_z(1-c_\theta) - a_x s_\theta & a_z^2 + c_\theta(1-a_z^2) \end{bmatrix}$$

where $a_x, a_y, a_z$ are the x , y and z coordinates of the axis of rotation $\vec{a}$. $c_\theta$ and $s_\theta$ are $\cos(\theta)$ and $\sin(\theta)$ correspondingly.

The constrained version of SUPR only limits the bones' degrees of freedom, by constraining the rotation matrices of the corresponding joints. Therefore,

this is an additional functionality, which we will release with SUPR, that can be enabled or disabled by a user of regression model.

## 8   Model Comparison

SUPR is trained on a federated dataset of head, body and head registrations. As a consequence of the sparse factorization of the pose space, we are able to separate the model into body part models. A comparison between SUPR and existing body models is shown in Figure 15.

| Model Name | Sparse Pose Deformations | Federated Training Data | Articulated Hands | Expressive Head | Part Based | Game Engine Compatibility | Publicly Available |
|---|---|---|---|---|---|---|---|
| SCAPE | ✗ | ✗ | ✗ | ✗ | ✗ | ✗ | ✓ |
| Stitched Puppet | ✓ | ✗ | ✗ | ✗ | ✓ | ✗ | ✓ |
| SMPL | ✗ | ✗ | ✗ | ✗ | ✓ | ✓ | ✓ |
| SMPL-H | ✗ | ✓ | ✓ | ✗ | ✗ | ✓ | ✓ |
| Frank | - | ✗ | ✓ | ✓ | ✗ | ✓ | ✓ |
| SMPL-X | ✗ | ✓ | ✓ | ✓ | ✗ | ✓ | ✓ |
| GHUM | ✗ | ✓ | ✓ | ✓ | ✗ | ✗ | ✓ |
| STAR | ✓ | ✗ | ✗ | ✗ | ✗ | ✓ | ✓ |
| BLSM | ✗ | ✗ | ✗ | ✗ | ✗ | ✓ | ✗ |
| **SUPR** | ✓ | ✓ | ✓ | ✓ | ✓ | ✓ | ✓ |

Fig. 15: A comparison between SUPR and existing body models.

### 8.1   SUPR

| Model | # Pose | # Joints | # Blendshapes |
|---|---|---|---|
| SUPR | 225 | 75 | 296 |
| SMPL-X [9] | 165 | 55 | 486 |
| GHUM [12] | 124 | 63 | - |

Table 3: **Body Models Comparison:** Comparing existing expressive human body models according to the number of pose parameters, number of joints and number of pose corrective blendshapes.

SUPR is a compact model that is compatible with the existing gaming and animation industry standards. The number of parameters of SUPR compared to existing expressive human body models is summarised in Table 3.

*Comparison with SMPL-X:* SUPR has 30% fewer pose-corrective blendshapes, despite having significantly more joints compared to SMPL-X. This is because of the Quaternion-based representation, which is significantly more compact compared to the Rodrigues representation used by SMPL-X. However, despite SUPR's compactness, it uniformly generalizes better than SMPL-X. The shape space of SMPL-X is trained on the CAESAR dataset [10], while SUPR is trained on $15,000$ registrations from both CAESAR and SizeUSA [1]. The SizeUSA dataset contains a larger diversity of body shapes and, in addition, the female subjects wore a traditional bra, whereas, in the CAESAR dataset, the female subjects wore a sports bra. The pose space of SMPL-X is trained on 2000 full body registrations. In contrast, SUPR's pose space is trained on a federated dataset of 1.2 million registrations of head, hand, body and feet registrations.

SMPL-X's pose blendshape formulation is based on SMPL. As a result, SMPL-X suffers from the same drawbacks of SMPL, namely SMPL-X also learns false long range spurious correlations; e.g. bending one elbow results in a bulge in the other elbow.

*Comparison with GHUM:* The GHUM model [12] pose space deformation function (PSD) is modeled by a neural network, which is not compatible with the gaming and animation industry standards. SUPR's learned blendshapes are linearly related to the model pose parameters, and hence the formulation is full compatible with the gaming and animation industry standards. While both SUPR and GHUM are trained on a federated dataset, and the GHUM authors propose a separated suite of models (GHUM-Head and GHUM-Hand), there are key important differences. The GHUM shape space is trained only on the CAESAR data (5K subjects),while SUPR shape space is trained on both CAESAR and SizeUSA, for a combined total of 15K registrations. On the other hand, the pose space of GHUM is trained on a dataset of 60K head, hand and body registrations, while the SUPR pose space is trained on 1.2 million body, head, hand and feet registrations. SUPR is the first to train on dedicated foot registrations. This is crucial for modeling realistic foot deformations due to movement of the ankle or curling of the toes, since the feet are consistently poorly reconstructed in full-body scans.

The GHUM PSD formulation is a dense non-linear formulation, where all the joints are related to all the vertices using a VAE [6]. As a result the body pose-space formulation of GHUM can not be separated into compact body parts. To define separate body part models, the GHUM authors segment the mesh and re-train the PSD function of the separated parts. The proposed head and hand models for GHUM fail to capture the full degrees of freedom of th head. SUPR and the separated head/hand models are jointly trained once. In contrast to GHUM, the SUPR pose-space formulation is strictly sparse, where each joint only influences a sparse set of the model vertices. As a result, SUPR can be separated into a suite of compact models. The learned kinematic tree of SUPR-Head has significantly more joints (neck and shoulders).

All prior expressive human body models ignore the human foot. The kinematic tree contains an additional 24 joints for modeling the full range of motion of the ankle and the toes.

## 8.2   SUPR-Head

The SUPR-Head has a pose, shape and expression space. We train 3 head models: female, male and a gender neutral model. The pose blendshape function is a subset of the learned SUPR pose corrective blendshapes, which are also sparse and spatially local. A comparison between and existing full head models is shown in Table 4.

| Model | # Pose | # Joints | # Blendshapes |
|---|---|---|---|
| SUPR-Head | 29 | 10 | 40 |
| FLAME [9] | 12 | 4 | 36 |
| GHUM-Head [12] | 23 | 10 | - |

Table 4: **Head Models Comparison:** Comparing existing head models models according to the number of pose parameters, number of joints and number of pose corrective blendshapes.

## 8.3   SUPR-Hand

We train a single gender-neutral SUPR-Hand model. SUPR-Hand has a pose and shape space. A comparison between SUPR-Hand and existing hand models is shown in Table 5. In comparison to MANO, SUPR-Hand has an additional wrist joint, which is necessary to model the hand deformations as a result of the wrist movement. A comparison between SUPR-Hand and existing hand models is shown in Table 5.

| Model | # Pose | # Joints | # Blendshapes |
|---|---|---|---|
| SUPR-Hand | 90 | 30 | 120 |
| MANO [9] | 90 | 30 | 270 |
| GHUM-Hand [12] | 18 | 36 | - |

Table 5: **Hand Models Comparison:** Comparing existing hand models according to the number of pose parameters, number of joints and number of pose corrective blendshapes.

### 8.4   SUPR-Foot

We train a male, female and neutral models for SUPR-Foot. SUPR-Foot is the first publicly-available articulated model of the human foot. We propose a novel deformation function that relates the foot deformation to the foot pose, foot shape and foot contact. SUPR-Foot shape space is trained on $7,000$ high-resolution foot scans that capture the diversity of the human foot shape variation. The pose space is trained on $57,231$ high-resolution scans that capture the foot sole deformations due to ground contact.

## 9   Limitation

A limitation of our method is that model training relies on registering a template mesh to the scans. While registration is automatic, the resulting data needs curation by an expert to detect any failures or artifacts. Registering 1.2M hand, head, body, and foot scans is time-consuming and labor-intensive. Registration remains a bottleneck for training body models on large datasets.

A key limitation when evaluating SMPL-X and GHUM against SUPR is that they are both trained on propriety data that is not publicly available to the research community. Additionally, the training code of SMPL-X and GHUM is also not publicly available for research purposes. Therefore, direct comparisons between the body models on the same data is difficult. Nevertheless, we are the first to evaluate all expressive body models on a publicly available test benchmark.


\end{document}